%% file: main.tex
\documentclass{article}

\PassOptionsToPackage{numbers, compress}{natbib}
\usepackage{graphicx}
\usepackage{wrapfig}

\usepackage[final]{neurips_2024}
\usepackage[utf8]{inputenc} % allow utf-8 input
\usepackage[T1]{fontenc}    % use 8-bit T1 fonts
\usepackage{hyperref}       % hyperlinks
\usepackage{url}            % simple URL typesetting
 
\usepackage{booktabs}       % professional-quality tables
\usepackage{amsfonts}       % blackboard math symbols
\usepackage{nicefrac}       % compact symbols for 1/2, etc.
\usepackage{microtype}      % microtypography
\usepackage{xcolor}         % colors
\usepackage{caption}
\usepackage{graphicx}
\usepackage{amsmath}
\usepackage{amssymb}
\usepackage{booktabs}
\usepackage{verbatim}
\usepackage{color}
\usepackage[english]{babel}
\usepackage{amsthm}
\theoremstyle{definition}
\newtheorem{definition}{Definition}[section]
\newtheorem{remark}[definition]{Remark}
\usepackage{float}
\usepackage{epsfig}
\usepackage{colortbl}
\usepackage{textcomp}
\usepackage{multirow}
\usepackage{algorithm}
\usepackage{algpseudocode}
\usepackage{caption}
\usepackage{caption,subcaption}
\usepackage{soul}
\usepackage[symbol]{footmisc}

\definecolor{customcolor}{RGB}{172, 44, 172}
\definecolor{customcolor2}{RGB}{16, 37, 240}
\definecolor{customcolor3}{RGB}{60, 80, 125}
\definecolor{customcolor4}{RGB}{64, 133, 167}
\definecolor{myblue}{rgb}{0.8,0.85,0.95}  % Light blue
\definecolor{myorange}{rgb}{1,0.8,0.6}  
\definecolor{blue}{HTML}{0066CC}
\definecolor{red}{HTML}{FF0201}
\definecolor{green}{HTML}{009900}
\usepackage{enumitem}
\usepackage{tikz}
\usepackage{annotate-equations}
\usepackage{mathtools}

\newcommand{\circled}[1]{\tikz[baseline=(char.base)]{
            \node[shape=circle,draw,inner sep=1pt] (char) {#1};}}

\title{\hspace{-0.4cm}
    \begin{minipage}{0.08\textwidth} % Adjust the width as needed for your image
        \includegraphics[width=\linewidth]{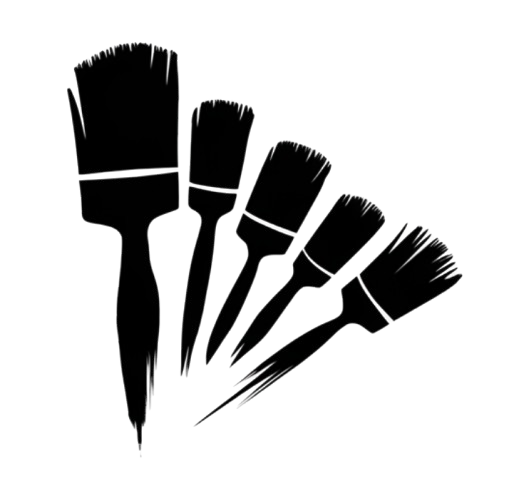} % Include your logo here
    \end{minipage}
    \hspace{-0.1cm}
    \begin{minipage}{0.92\textwidth} % The rest of the title text on the right
        \textbf{ParallelEdits: Efficient Multi-object Image Editing}
    \end{minipage}
}
\author{%
  Mingzhen Huang, Jialing Cai, Shan Jia, Vishnu Suresh Lokhande$*$, Siwei Lyu$*$ \\
  University at Buffalo, State University of New York, USA\\
}

\begin{document}
\input{definitions}
% \definecolor{mygray}{gray}{.9}

% TODO REVIEW: If the paper title is too long for the running head, you can set
% an abbreviated paper title here. If not, comment out.

\renewcommand\twocolumn [1] []{#1} \maketitle %\begin{center}
\vspace{-0.35in} 

\begin{figure}[!h] 
\captionsetup{type=figure} 
\includegraphics[width=\linewidth]{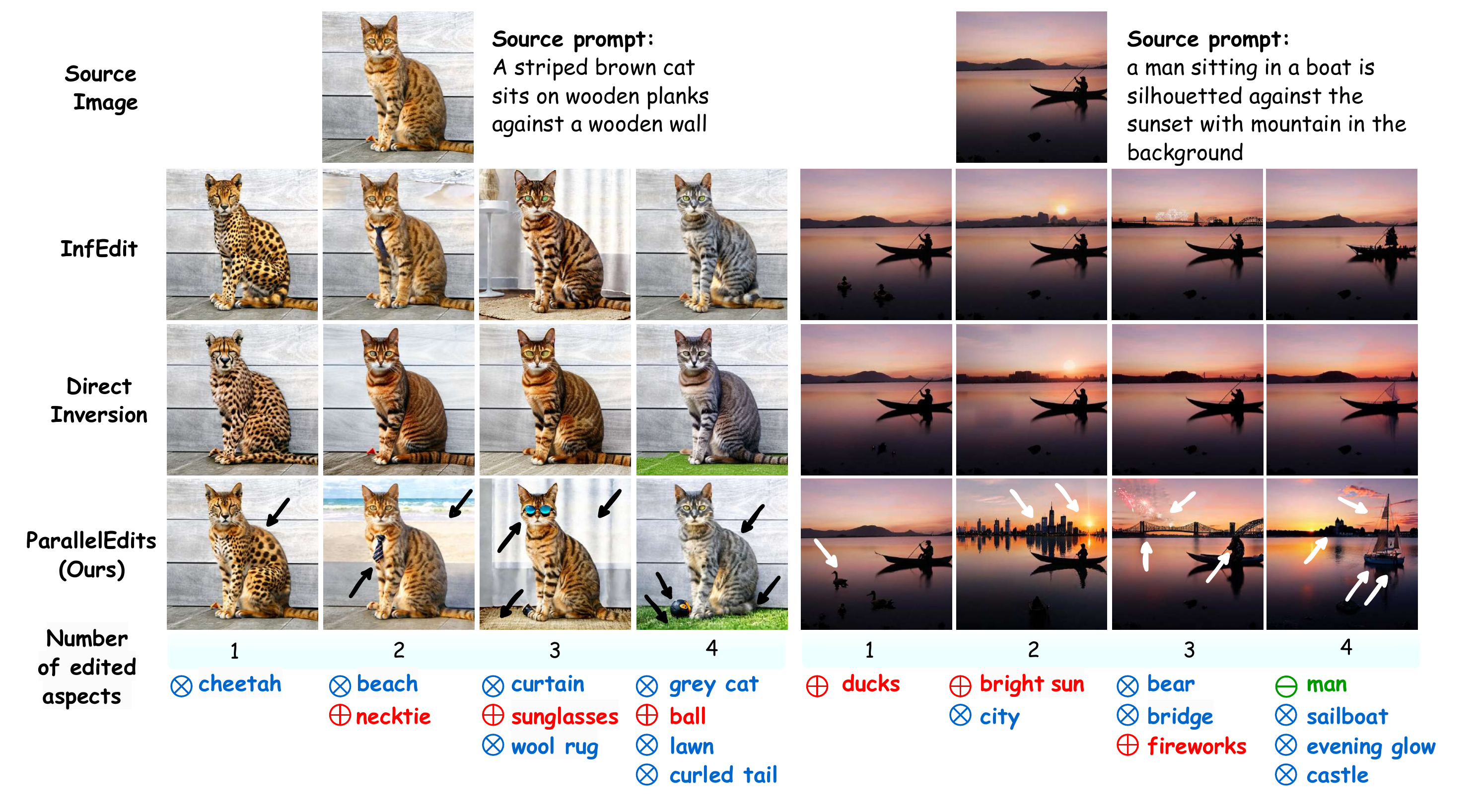}{
% \centering

\vspace{-0.15in}
         \caption{\footnotesize {\textbf{Multi-aspect text-driven image editing.} Multiple edits in images pose a significant challenge in existing models (such as DirectInverison \cite{ju2023direct} and InfEdit \cite{xu2023inversion}), as their performance downgrades with an increasing number of aspects. In contrast, our ParallelEdits can achieve precise multi-aspect image editing in 5 seconds}. The symbol $\textcolor{blue}{\boldsymbol{\otimes}}$ denotes a \textcolor{blue}{swap} action, the symbol $\textcolor{red}{\boldsymbol{\oplus}}$ denotes an object \textcolor{red}{addition} action, and the symbol $\textcolor{green}{\boldsymbol{\ominus}}$ denotes an object \textcolor{green}{deletion}. Arrows ($\rightarrow$) on the image highlight the aspects edited by our method. }\label{fig:teaser} }
% \label{fig:dataset}\

\vspace{-0.13in}
\end{figure} 

% \begin{figure}[]
%     %\vspace{-1cm}
%     \centering
%     \includegraphics[width=1.0\linewidth]{figs/teaser_jl.pdf}
%   %\vspace{-1.5em}
%     \caption{\small \em {Multi-aspect text-driven image editing poses a challenge for existing editing models, including DirectInverison \cite{ju2023direct}, InfEdit \cite{xu2023inversion}, and NTI \cite{mokady2023null} models, while our ParallelEdits can achieve precise multi-aspect image editing in 5 seconds}. We denote the \textcolor{red}{add} aspects as \textcolor{red}{red}, \textcolor{blue}{swap} aspects as \textcolor{blue}{blue} and \textcolor{green}{remove} aspects as \textcolor{green}{green} ({TO BE UPDATED!)}}
%     \label{fig:teaser}
%     \vspace{-0.45cm}
%   \end{figure}
  
\maketitle 
\footnotetext[1]{Corresponding authors}

% \vspace{-0.3in} 
% \captionof{figure}{\textbf{Output examples of proposed DETracker.} The box in \textcolor{yellow}{yellow} is the patch memory. The space in \textcolor{gray}{gray} and black indicates the compensated ego motion. } \label{fig:teaser}
%\end {center} 

% \maketitle
\begin{abstract}
    Text-driven image synthesis has made significant advancements with the development of diffusion models, transforming how visual content is generated from text prompts. Despite these advances, text-driven image editing, a key area in computer graphics, faces unique challenges. A major challenge is
    % maintaining consistency in the edits while ensuring the background remains unaffected, and there is no unintended alteration to areas not meant to be edited. While existing methods can produce consistent edits and maintain background integrity, they struggle with 
    making simultaneous edits across multiple objects or attributes. 
    Applying these methods sequentially for multi-aspect edits increases computational demands and efficiency losses. 
    In this paper, we address these challenges with significant contributions. Our main contribution is the development of ParallelEdits, a method that seamlessly manages simultaneous edits across multiple attributes. In contrast to previous approaches, ParallelEdits not only preserves the quality of single attribute edits but also significantly improves the performance of multitasking edits. This is achieved through innovative attention distribution mechanism and multi-branch design that operates across several processing heads. 
    % Crucially, the computational complexity of ParallelEdits does not increase with the number of heads, ensuring the method is both efficient and scalable. 
    Additionally, we introduce the PIE-Bench++ dataset, an expansion of the original PIE-Bench dataset, to better support evaluating image-editing tasks involving multiple objects and attributes simultaneously. 
    This dataset is a benchmark for evaluating text-driven image editing methods in multifaceted scenarios. Codes are available at: \href{https://mingzhen-huang.github.io/projects/ParallelEdits.html}{https://mingzhen-huang.github.io/projects/ParallelEdits.html}.
\end{abstract}

\section{Introduction}
\label{sec:intro}
% poses a challenge for existing editing models, including DirectInverison \cite{ju2023direct}, InfEdit \cite{xu2023inversion}, P2P \cite{hertz2022prompt}, and NTI \cite{mokady2023null} models, while our ParallelEdits can achieve precise multi-aspect image editing in 5 seconds}. We denote the \textcolor{red}{add} aspects as \textcolor{red}{red}, \textcolor{blue}{swap} aspects as \textcolor{blue}{blue} and \textcolor{green}{remove} aspects as \textcolor{green}{green}.
Recently, text-driven image editing has experienced remarkable growth, driven by advances in diffusion-based image generative models. This technique involves modifying existing images based on textual prompts to alter objects, their attributes, and the relationships among various objects. The latest methods ~\cite{hertz2022prompt, ju2023direct, cao_2023_masactrl} can produce edited images that closely match the semantic content described in the prompts while keeping the rest of the image unchanged. Unlike early image editing approaches that required image matting to precisely extract foreground objects using alpha mattes ~\cite{li2023referring}, text-driven editing offers a less labor-intensive alternative. User-provided textual prompts guide the edits, with auxiliary inputs like masks facilitating localized modifications~\cite{liu2021tripartite}.
% With a wide range of applications from entertainment to e-commerce advertising and metaverse platforms, text-driven image editing has attracted significant interest from the research community~\cite{sheynin2023emu,li2023deep}.

%{TO BE ADDED: 1) Why multi-aspect editing is necessary. 2) High-level definitation of multi-aspect editing. 3) How it is different from single-aspect editing and full synthesis. ours task holds a broader range and is a bridge between these two, high flexibility, a moving goal)}. 

%While these methods have showcased promising results, existing methods mainly assume a single aspect in the source image (\eg, object type, object attribute, and object relation) needs to be edited. The task of altering multiple aspects of an image simultaneously through a text prompt, a problem we term {\em multi-aspect text-driven image editing}, is rarely explored. %This is understandable, as they were mostly tested with text prompts that only change one aspect of the image content. 
%However, the quality and performance degrade significantly when the user intends to change multiple aspects of an image {\em simultaneously} through the text prompt, a problem we term {\em multi-aspect text-driven image editing}. %Fig.\ref{} (a). 
While these methods have showcased promising results, existing methods typically focus on editing a single aspect in the source image. An ``aspect'' refers to a specific attribute or entity within the textual prompt that describes the image and can be modified, such as object type, color, material, pose, or relationship. However, the ability to edit multiple aspects through text prompts is rarely explored. We introduce the concept of {\em multi-aspect text-driven image editing} to address this gap. Multi-aspect image editing is essential due to the rich and diverse content and structure of digital images, as well as the varied requirements of users. For example, it always occurs that users wish to modify multiple attributes or regions in an image, such as adding a necktie to a cat and changing the background wall to a beach (Fig. \ref{fig:teaser}, Left), or removing a man and replacing a mountain with a castle in the right example. Unlike traditional editing methods (e.g., \cite{ju2023direct, xu2023inversion}) that focus on a single aspect, multi-aspect editing allows users to manipulate various aspects simultaneously. %, such as attribute changes with object addition or deletion. 
Different from full text-to-image synthesis~\cite{gu2022vector, ding2022cogview2}, which involves creating content from scratch, multi-aspect editing works with the source image to ensure essential content preservation. It bridges the gap between single-aspect editing and full synthesis, catering to a wide range of editing scenarios.

However, we observe that directly applying the single-aspect text-driven image editing methods in cases where multiple image aspects must be modified often does not yield satisfactory results. A straightforward solution to this problem is to apply the single aspect editing method {\em sequentially} -- we can order the aspects to be modified and use a single-aspect editing method to change the aspects one by one. Although sequential applications of single-aspect text-driven image editing methods can modify multiple aspects of an image, they may introduce significantly higher computational overhead. More importantly, the order of the aspects modified may affect the quality -- changes to later aspects may undo the early ones or accumulate the errors and artifacts, thus reducing the effectiveness of the final editing results, as the last two rows of \Fref{fig:comapre} and \Tref{tab:main} show. %, as Fig.\ref{} (b) shows.

In this work, we introduce {\em ParallelEdits} as an efficient and effective solution to the problem of multi-aspect text-driven image editing. This method is based on a crucial insight that the editing step can occur in parallel with the image's diffusion steps. Therefore, in ParallelEdits, we build image aspect editing into the diffusion steps to accelerate the editing process. ParrallelEdits is based on an architecture with a fixed number of additional branches dedicated to handling rigid, non-rigid, and style changes. This design ensures scalability independent of the number of prompt aspects altered. In addition, we employ an attention aggregator to accurately assess editing difficulty and route aspects to appropriate branches within the ParallelEdits framework, ensuring precise and efficient editing. To enable subsequent research and evaluation of multi-aspect text-driven image editing methods, we also build the PIE-Bench++ dataset, an extension of the PIE-Bench~\cite{ju2023direct} that has $700$ images with detailed text prompts and tailored to facilitate simultaneous edits across multiple image aspects. We propose evaluation metrics and benchmark different text-driven image editing methods on PIE-Bench++. The ParallelEdits outperforms the state-of-the-art image editing methods on PIE-Bench++. 

%This improvement is achieved through a novel attention distribution mechanism distributed over multiple heads. Importantly, t

% \end{itemize}

% \begin{figure}[!t]
%     \centering
%    \includegraphics[width=0.65\textwidth]{figs/teaser.pdf}
     
% %     % \vskip -0.15in
%      \label{fig:teaser2}
%      \end{figure}

\section{Related Works}
%\VISHNU{Can we add this one too? Like how we are different from multi-view editing \url{/https://arxiv.org/pdf/2402.14792.pdf}}

%\VISHNU{PLEASE MERGE THE FOLLOWING TWO POINTS INTO THE RELATED WORKS TEXT}

%{\color{red} Why single edits method did not work well for multi-attributes?Edits method relies on attention mask for local region edit, for multiple attributes, the edit region mask is huge and have more semantic information in it for editing
%There are two kinds of single att edits method: 1) attention mask is fix for cross attention which does not work well for pose-changed obj 2) adaptive mask that can be used for pose-changed obj edit but it is less accurate. So how to combine both two edit method for multi-attribute is a technical problem. }

%The integration of Generative Adversarial Networks (GANs) with the Contrastive Language-Image Pre-training (CLIP) framework has advanced text-based image manipulation, enabling the generation of high-quality images~\cite{radford2021learning}. However, GAN-based approaches face challenges when confronted with diverse and large datasets~\cite{zhu2023exploring}. 

%that are built from Trimaps~\cite{liu2021tripartite}, scribbles~\cite{yang2020smart}, or text descriptions~\cite{li2023referring} due to the inherently ill-posed nature of the task. The latest advancements in text-driven image editing methods have placed a significant emphasis on diffusion steps as their core mechanism, exemplified by prominent techniques such as InfEdit \cite{xu2023inversion}, MasaCtrl~\cite{cao_2023_masactrl}, and Prompt-to-Prompt (P2P) \cite{hertz2022prompt}. 

\label{sec:rel-works}
\myheading{Diffusion Models for Text-Driven Image Editing}. Text-driven image editing aims to manipulate local regions of an image based on textual prompts. The editing has two main goals: ensuring the edits align with provided instructions and preserving essential content. Diffusion models~\cite{rombach2022high} have gained popularity as a preferred image editing model for their capacity for generating high-quality samples by incorporating diverse conditions, especially using text \cite{couairon2023diffedit, parmar2023zero, xu2023inversion, kawar2023imagic, couairon2022diffedit,nguyen2024visual,ju2023direct}. %, image \cite{yang2023paint, gu2024photoswap}, and tactile sensing\cite{yang2023generating}.
This involves transforming the images into the latent space and generating regions using diffusion models conditioned by the text prompt while ensuring accurate reconstruction of unmodified regions during editing. To avoid the edited image deviating from original image, early text-driven image editing typically requires user-specified masks as additional condition~\cite{lugmayr2022repaint,avrahami2022blended, nichol2021glide} or training~\cite{choi2021ilvr, xu2024cyclenet, zhao2022egsde} to guided the editing process, which constrain their potential zero-shot application. To address this limitation, recent editing models, such as InfEdit \cite{xu2023inversion}, PnP~\cite{tumanyan2023plug}, Direct Inversion~\cite{ju2023direct} follow the work Prompt-to-Prompt (P2P) \cite{hertz2022prompt}, which proposed to obtain an attention map from the cross attention process and either swap or refine the attention map from text prompt for image editing. This design automatically obtains the editing mask and only allows image editing using a text prompt. Another method, MasaCtrl~\cite{cao_2023_masactrl}, converts existing self-attention in diffusion models into mutual self-attention for non-rigid consistent image synthesis and editing, enabling to query correlated local contents and textures from source images for consistency.  

\myheading{{Multi-Aspect Image Editing}}.
While current image editing models have shown promising results in their text-driven image editing benchmarks, we have observed that they work well on single-attribute editing while struggling to edit multiple aspects, especially when editing multiple objects (as shown in \Fref{fig:teaser}). %This is limited by the simple attention map strategy for editing mask learning. Why single edits method did not work well for multi-attributes?  
We attribute this limitation to the following reasons. First, existing methods use the attention mask to direct where edits should be made. With multiple attributes, the editing area may expand significantly, incorporating extensive semantic information or scattered regions that are challenging to edit using a single mask. Second, employing a fixed mask from cross-attention maps struggles with edits involving changes in region size (such as pose adjustments), while using an adaptive mask faces challenges in maintaining edit fidelity. Therefore, integrating various attention masks for accurate multi-attribute editing presents a challenging technical problem.
%There are two kinds of single att edits method: 1) attention mask is fix for cross attention which does not work well for pose-changed obj 2) adaptive mask that can be used for pose-changed obj edit but it is less accurate. So how to combine both two edit method for multi-attribute is a technical problem.gi
%\myheading{Multi-aspect Image Editing.} 
Early studies~\cite{wang2022maniclip, khodadadeh2022latent} have employed GAN models such as StyleGAN2~\cite{karras2020analyzing} to edit multiple attributes in faces. The multiple-attribute editing is realized by training the GAN model with supervised multi-class training and a training dataset of image and attribute vector pairs. This solution heavily relies on the training sets and has limitations in generalizing to new editing types. {Few recent works achieve multi-aspect editing with additional inputs: \cite{ge2023expressive} leverages rich text to edit multiple objects and ~\cite{chang2024ground} pre-processes the image with grounding to localize multiple edited regions for multi-aspect editing. However, the editing performance highly relies on additional input beyond plain text, either from user input or other off-the-shelf models.} A recent work \cite{joseph2024iterative} proposes an iterative multi-granular image editor, where a diffusion model can faithfully follow a series of image editing instructions from a user. However, this interactive editing pipeline will result in significant computational overhead.% and reduced runtime efficiency. 
%\citet{guo2023focus} proposed a fine-grained and multi-instruction image editing by attention modulation. Different with instruction-based image editing that take an instruction as input (e.g., turn the red dogs to blue), text-driven image editing  directly edit the image based on the text prompt without the aspects correspondence from from user input.
%Face related: \cite{wang2022maniclip}

%Comparing with editing single aspesct, their performance downgrades a lot when editing multiple aspest especially on editing multiple objects as shown in \Tref{}.

\myheading{Image Editing with Multiple Branches.} In the literature \cite{cao_2023_masactrl, hertz2022prompt}, image editing processes have been conducted by implementing a dual-branch approach. This methodology involves segregating source and target branches throughout the editing process. Specifically, the source branch is reverted to $z_0$, while the trajectory of the target branch is iteratively adjusted. By computing the distance from the source branch, the calibration of the target branch occurs at each time-step.
Our observation underscores the disparity between the effectiveness of a dual branch in enhancing the editing process and its failure in multi-aspect editing. A singular target branch proves inadequate in calibrating fully from the source branch, leading to imperfect incorporation of all aspects into the image. Hence, our primary proposition advocates for multi-aspect editing by utilizing multiple target branches. Each target branch's trajectory is meticulously calibrated, with simpler concepts addressed in the initial branches and more complex aspects deferred to subsequent ones. In the following section, we will delve deeper into this concept.
%\VISHNU{PLEASE ADD EQUATIONS 7-12 from ALGORITHM 2 of INFEDIT OVER HERE. ALTERNATIVELY, INSTEAD OF EQUATIONS HAVING IT AS FIGURE 2 of \url{https://arxiv.org/pdf/2402.13490v1.pdf} IS ALSO OKAY}
\section{Diffusion-based Image Generation and Editing}
\label{sec:prelim}
We are provided with an image sample  $x_0$  which transforms the latent space via an encoder/decoder pair \(\mathcal{E}/\mathcal{D}\), such that \( z_0 = \mathcal{E}(x_0) \). Here, \( z_0 \) represents the latent representation of the image \( x_0 \). With a slight abuse of notation, we approximate the reconstructed image \( \bar{x}_0 \) as \( \mathcal{D}(\bar{z}_0) \), where \( \bar{z}_0 \) denotes the reconstructed version of \( z_0 \). These operations are integral to the latent diffusion model \cite{rombach2022high}. The diffusion process constitutes two steps: the forward step incrementally adds zero-mean white Gaussian noise with time-varying variance to the latent vector $z$ according to discrete-time $t$\footnote{Diffusion process is rigorously defined as a continuous-time stochastic differential equation, but in practice often implemented with discrete-time updates.},
\begin{align}
    z_t = \sqrt{\alpha_t} z_0 + \sqrt{1-\alpha_t} \epsilon \quad \text{with} \quad \epsilon \sim \mathcal{N}(0, I),
\end{align}
$\alpha_{1:T}$ represents a variance schedule for $t$ drawn from the interval $[1,T]$. The variance schedule can be different, such as linear or cosine quadratic \cite{nichol2021improved}. The backward step is an iterative process to remove the noise from the data progressively. Using the same variance schedule $\alpha_{1:T}$ as in the forward step, a noise schedule $\sigma_{1:T}$ and a parameterized noise prediction network $\epsilon_\theta$ with coefficients $c_{\text{pred}} = \sqrt{\alpha_{t-1}}$, $c_{\text{dir}} = \sqrt{1-\alpha_{t-1}-\sigma^2_t}$, and $c_\text{noise}=\sigma_t$, the backward step corresponds to the following process:
\begin{align}
    z_{t-1} = \underbrace{c_{\text{pred}} f_\theta(z_t, t)}_{\text{predicting $\bar z_0$}} + \underbrace{c_{\text{dir}} \epsilon_\theta(z_t, t)}_\text{adjust along $z_t$} +  \underbrace{c_\text{noise}\epsilon_t}_\text{random noise} \quad \text{with} \quad \epsilon_t \sim \mathcal{N}(0, I)
\end{align}
The noise schedule \( \sigma_{1:T} \) comprises hyperparameters requiring careful selection based on factors like image dimensions or desired performance \cite{chen2023importance}\cite{karras2022elucidating}. In the framework of Denoising Diffusion Implicit Models (DDIM) \cite{song2020denoising}, the function \( f_\theta \) is employed for the prediction and reconstruction of \( \bar z_0 \), based on the input \( z_t \). Specifically, we have $\bar z_0 = f_\theta(z_t, t) = \frac{1}{\sqrt{\alpha_t}}z_t - \frac{\sqrt{1-\alpha_t}}{\sqrt{\alpha_t}} \epsilon_\theta(z_t, t)$.

\myheading{Consistency Models for Inversion-free Image Editing}. 
Consistency models~\cite{song2023consistency, luo2023latent} have been introduced to expedite the generation process through a consistent distillation approach. These models exhibit a self-consistency property, ensuring that samples along the same trajectory map to the same initial point. Specifically, the function $f_\theta$ is rendered self-consistent by satisfying $f_\theta(z_t, t) = z_0$ for a given sample $z_t$ at timestep $t$. As a result, the self-consistency property yields a closed-form solution for the noise predictor $\epsilon_\theta$. We denote this particular $\epsilon_\theta$ as $\epsilon^\text{cons}$, which is derived as $\epsilon^\text{cons} = \frac{z_t-\sqrt{\alpha_t}z_0}{\sqrt{1-\alpha_t}}$. Since $\epsilon^\text{cons}$ is not parameterized and contains the ground-truth $z_0$, Xu \textit{et al.}~\cite{xu2023inversion} propose starting directly with random noise, i.e., $z_T \sim \mathcal{N}(0, \mathbf{I})$, at the last time-step $T$, which is particularly advantageous for image-editing tasks as it eliminates the need for inversion from $z_0$ to $z_T$. Therefore, starting with $z_\tau = z_T \sim \mathcal{N}(0, \mathbf{I})$, the sampling process proceeds as follows:
\begin{enumerate}[label={\protect\circled{\arabic*}},itemindent=0pt,leftmargin=*,labelsep=1em,itemsep=0pt,parsep=0pt,topsep=0pt]
    \item $z = \frac{z_\tau - \sqrt{1-\alpha_\tau}\epsilon_\tau^{\text{cons}}}{\sqrt{\alpha_\tau}}$. Where,  $\epsilon_\tau^{\text{cons}}$ is given by $\frac{z_\tau-\sqrt{\alpha_t}z_0}{\sqrt{1-\alpha_t}}$ \label{item:first_step}
    \item Noise is added to $z_\tau$, i.e, $z_\tau= \sqrt{\alpha_\tau}z + \sqrt{1-\alpha_\tau}\epsilon$ where $\epsilon\sim \mathcal{N}(0, \mathbf{I})$ \label{item:second_step}
\end{enumerate}
After many iterations, the final output is $z$. Furthermore, \cite{xu2023inversion} demonstrates that the dual-branch paradigm (involving a source and a target branch) used in image editing tasks can be executed in an inversion-free manner. We will delve into this, along with our method description, in Section~\ref{sec:multi_branch}.

\section{Multi-Aspect Image Editing}
\label{sec:method}  

%% Dataset Lookup 
% Show in the figure how the attention map will look like  

%% DRY RUN

%% SMART ATTENTION LOOKUP

% In the paper, we proposed a diffusion-based training-free method, ParallelEdits, to handle multi-aspect text-driven image editing. 
% The method is based on a novel design of attention distribution, each local region of the input image is separately edited by a
% different branch. Besides that, there are three different type of branches in proposed pipeline that aim to handle different type of local/global region edits:  
% rigid editing (RE), non-rigid editing (NRE) and global style adaptive editing (SA).
% Those three different branches work parallel and the input is concatenated in the batch size dimension, thus the denoising UNet~\cite{ronneberger2015u} of Diffuision Models~\cite{rombach2022high} is only forwarded once. 
% Along with the virtual inversion technical proposed in \cite{xu2023inversion}, 
% our proposed method can achieve multi-aspect image edit in around 5 seconds. 
\subsection{Problem Definition}
% TODO

The input to the multi-aspect image editing task includes a source image ($\mI_{src}$), the source prompt, and a set of edits to be applied to the source image, indicating the changes from the source prompt to target prompt. A text prompt (whether source or target) comprises several independent tokens, of which only a subset is editable. We refer to these editable tokens as \textit{Aspects}.
\begin{definition}[Aspect] {We define an $i^{\text{th}}$ aspect $\mA_{src}^i$ in the source prompt (or the $j^{\text{th}}$ aspect $\mA_{edt}^j$ in the target prompt) as any entity that can be substituted, deleted, or inserted into the text prompt, resulting in a meaningful sentence structure.}
\end{definition}
Several examples of tokens corresponding to aspects or not are given in Fig.~\ref{fig:attn}. In other words, aspects correspond to single or multiple tokens representing object color, pose, material, content, background, image style, etc. An editing operation $E^{i \rightarrow j}$ between the editing pair $(\mA_{src}^i, \mA_{edt}^j)$ as $E^{i \rightarrow j} \in \{\textcolor{blue}{\boldsymbol{\otimes}}, \textcolor{red}{\boldsymbol{\oplus}}, \textcolor{green}{\boldsymbol{\ominus}}, \textcolor{black}{\boldsymbol{\oslash}}\}$. Here, $\textcolor{blue}{\boldsymbol{\otimes}}$ denotes a swap action, $\textcolor{red}{\boldsymbol{\oplus}}$ denotes an object addition action, $\textcolor{green}{\boldsymbol{\ominus}}$ denotes object deletion, and $\textcolor{black}{\boldsymbol{\oslash}}$ indicates no change in the aspect. Such an editing operation can be inferred directly by appropriately mapping the source and target prompts, or it can be provided as metadata \cite{hertz2022prompt, mokady2023null}.
\iffalse
\begin{remark}
{A text-driven image editing tasks involving $k$-aspects \footnote{Unlike many popular image-editing tasks \cite{ju2023direct, xu2023inversion}, our multi-aspect editing framework encourages $k$ to be larger than 
$1$. We evaluate the method based on its ability to perform multiple edits for $k \ge 1$.} can be modeled as,  $\mI_{edt}=Edit(\mI_{src},\mA_{{edt}_1}^{i_1 \rightarrow j_1},\mA_{{edt}_2}^{i_2 \rightarrow j_2},...,\mA_{{edt}_k}^{i_k \rightarrow j_k})$, where $k \geq 1$  and $E^{i \rightarrow j}  \not = \textcolor{black}{\boldsymbol{\oslash}}$.}
\end{remark}
Therefore, it is evident that the effectiveness of a multi-aspect editing method is determined by its ability to successfully execute editing operations $E^{i \rightarrow j}  \notin \{\textcolor{black}{\boldsymbol{\oslash}}$\} whilst preserving the $\oslash$ editing operation.
\fi
The editing task is considered successful if the edited source image, $\mI_{edt}$, reflects the required edits while preserving the unaffected aspects of the original image.

\begin{figure}[ht]
    \centering
    \includegraphics[width=1.0\linewidth]{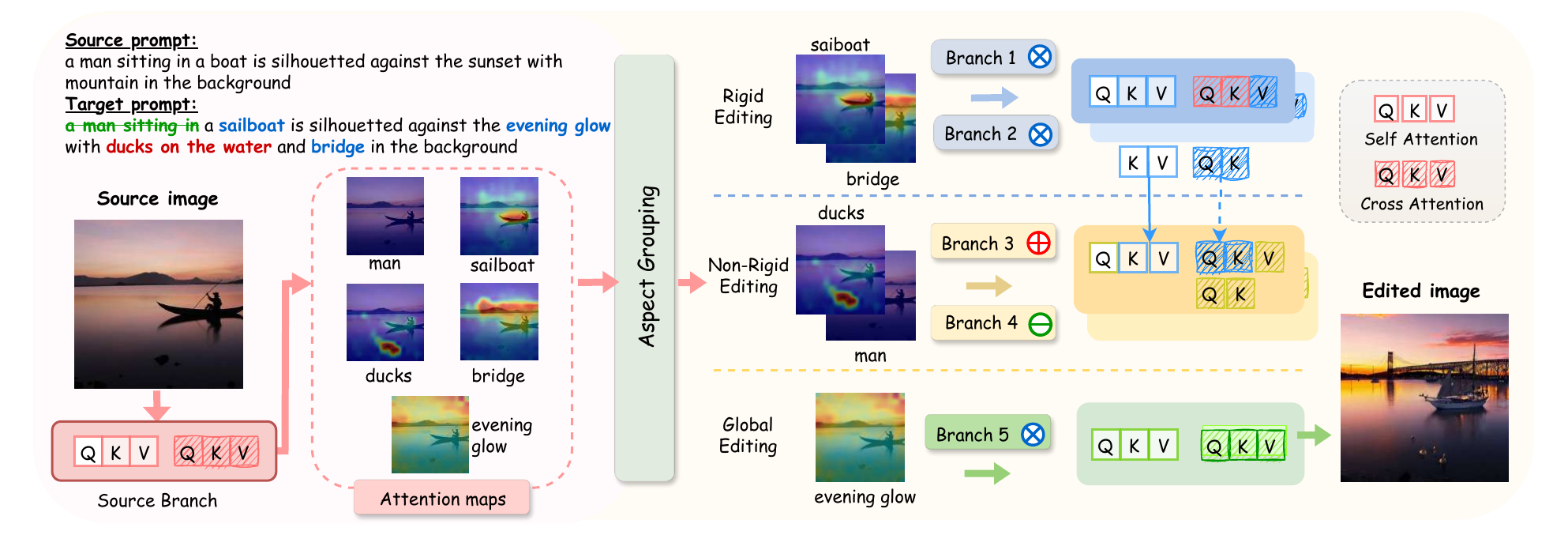}
  \vspace{-1.5em}
    % \caption{\footnotesize \textbf{Overall pipeline of ParallelEdits.} Added aspects are noted in \textcolor{red}{red}, swapped aspects are noted in \textcolor{blue}{blue}, and those aspects are grouped by attention maps while the editing for each group would be conducted in an individual branch.}
    \captionsetup{font=small} 
    \caption{\textbf{Pipeline.} Our method, ParallelEdits, takes a source image, source prompt, and target prompt as input and produces an edited image. The target prompt specifies the edits needed in the source image. Attention maps for all edited aspects are first collected. Aspect Grouping (see Section~\ref{sec:aspect_group}) categorizes each aspect into one of $N$ groups (in the above figure, $N=5$). Each group is then assigned a branch and the branch-level updates are detailed in Section~\ref{sec:multi_branch}. Each branch can be viewed either as a rigid editing branch, non-rigid editing branch, or global editing branch. Finally, adjustments to query/key/value at the self-attention and cross-attention layers are made, as illustrated in the figure and described in Section~\ref{sec:cross-branch}.}
    \label{fig:pipeline}
    \vspace{-0.45cm}
  \end{figure}

\subsection{Method}
Figure~\ref{fig:pipeline} outlines the overall pipeline of our method, which has three steps. In the first step (\Sref{sec:aspect_group}), we perform {\em aspect grouping} using attention maps generated by running a few iterations of the diffusion process. The aspects in the source image are put into up to $N$ groups, each processed by a distinct branch. The second step (\Sref{sec:multi_branch}) demonstrates how each branch, which receives a specific group of aspects, performs inversion-free editing. In the last step (\Sref{sec:cross-branch}), we perform the necessary adjustments for enabling cross-branch interaction and elucidate the significance of such interaction.

%\new{To achieve the multi-aspect editing task, our method's key idea is to group multiple edits and then perform the editing using our proposed multi-branch diffusion model. Specifically, we first categorize and group the editing by their attention maps. Then, we initialize multiple branches in the diffusion model based on the number of groups. Finally, we propose to use Consistency Models~\cite{song2023consistency} for multi-branch diffusion inversion and assign each grouped edit to a branch for the image editing and obtain the final target image.}

\subsubsection{Aspect Grouping}
\label{sec:aspect_group}
% {\color{red} https://arxiv.org/abs/2210.04885
% Attentiond maps define the layout of the image https://textual-inversion.github.io/ 

% - Normalization of attention maps 

% - size  of the attention of amps 
% }
\begin{wrapfigure}{!h}{0.5\textwidth}
\begin{center}
\vspace{-2em}
\includegraphics[width=0.48\textwidth]
{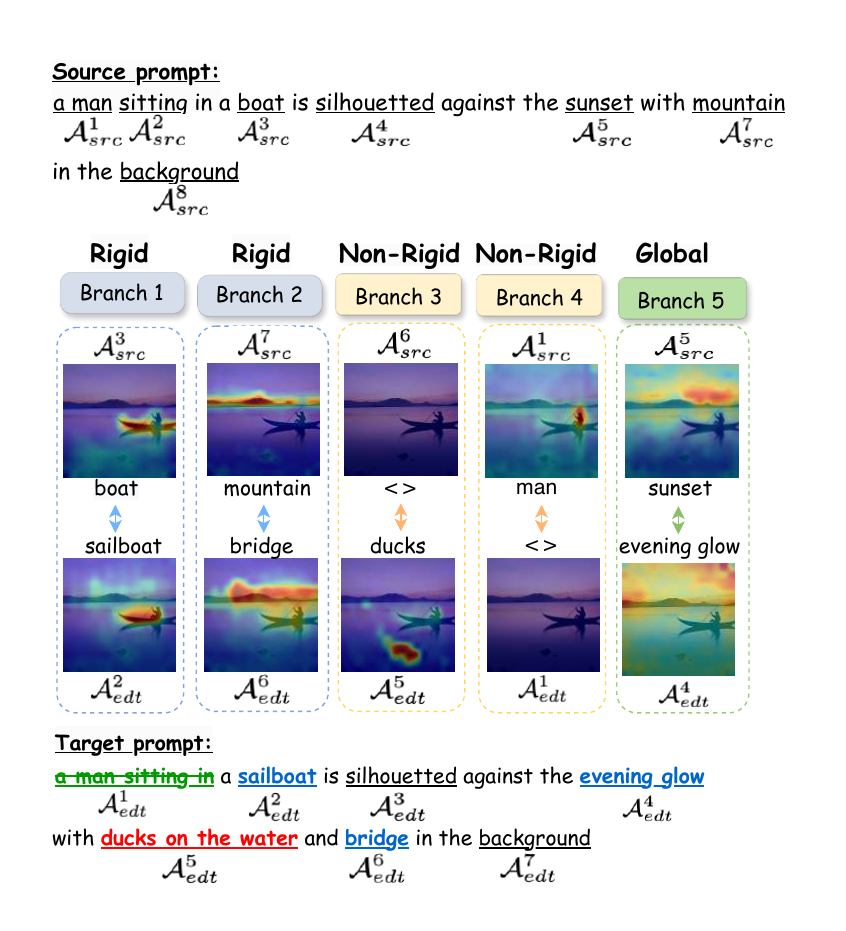}%Fig1_last.pdf}
\vspace{-0.5em}
% \caption{\footnotesize \textbf{Aspect grouping.} We group aspect pairs from the source prompt to target prompt into different branches based on their attention maps.}
\caption{\footnotesize \textbf{Aspects and Aspect Grouping.} In a text prompt, there are multiple independent tokens, with only some being editable, known as aspects and are underlined in the above example. These aspects can be added, deleted, or swapped between the source and target prompts. Pairs of source and target aspects are grouped into branches, and the methodology for aspect grouping is explained in Section~\ref{sec:aspect_group}.}
\label{fig:attn}
\vspace{-3em}
\end{center}
\end{wrapfigure}
We would like to group aspects in a prompt into \( N \) distinct groups using the cross-attention maps of the diffusion UNet \cite{ronneberger2015u} to characterize the spatial layouts as in previous studies \cite{tang2022daam}. Given an editing operation \(E^{i \rightarrow j}\) between the source aspect \(\mA_{src}^i\) and the target aspect \(\mA_{edt}^j\), we obtain the corresponding attention maps from both the source and target prompts as \(\bar{\mM}_{src}^i\) and \(\bar{\mM}_{edt}^j\), respectively. The attention map \(\mM\) is defined by the query feature \(\hat{Q}\) and key feature \(\hat{K}\) from the cross-attention as \(\mM = \text{softmax}\left(\frac{\hat{Q}\hat{K}^T}{\sqrt{d}}\right)\). The binarized attention map \(\bar{\mM}\) is obtained by normalizing \(\mM\) and thresholding its values. Our aspect grouping proceeds in two steps, \\
\vspace{0.1cm}
{\bf Step 1. Assign a type for every editing operation $\mathbf{(E^{i \rightarrow j})}$. }
We consider three possible types of edits, in line with previous works \cite{cao_2023_masactrl}, namely a global edit, a local rigid edit or a local non-rigid edit. Rigid local edits, such as changing an object's color or texture, do not alter the layout of objects. Conversely, non-rigid local edits modify the layout of objects, such as adding or deleting objects or changing object poses. Global edits affect background and style changes. The type assignment for the editing operation $(E^{i \rightarrow j})$ is determined by the following rules:

{\small
\begin{equation}
\label{eq:edit_type}
    \text{type}(E^{i \rightarrow j}) = \begin{cases} 
    \text{global edit} \quad.......................................................\; \gamma(\bar{\mM}_{edt}^j) \geq \beta \gamma\left(\sum \{\bar{\mM}_{edt}\}\right) \\
    \begin{rcases}
        \text{non-rigid edit} & \phi(\bar{\mM}_{src}^i, \bar{\mM}_{edt}^j) < \lambda \\
        \text{rigid edit} & \phi(\bar{\mM}_{src}^i, \bar{\mM}_{edt}^j) \geq \lambda
    \end{rcases} \text{local edit}\; ... \;\gamma(\bar{\mM}_{edt}^j) < \beta \gamma\left(\sum \{\bar{\mM}_{edt}\}\right)
    \end{cases}
\end{equation}
}

Here, $\phi$ represent mIoU~\cite{lin2014microsoft}, while $\gamma$ indicates the alpha mattes of attention maps. $\lambda$ and $\beta$ are tunable hyperparameters. For further details, please refer to the supplementary \Sref{sec:impl}. 

{\bf Step 2. Categorize every editing operation $\mathbf{(E^{i \rightarrow j})}$ into $\mathbf{N}$ groups. }
For each editing operation $(E^{i \rightarrow j})$ of a specific type, we assess whether $\phi(\bar{\mM}^j_{edt}, \bar{\mM}^k_{edt}) \geq \lambda$ to determine if there exists substantial overlap between any pair of attention maps of that type. If significant overlap is detected, the attention maps are grouped together. On the other hand, if attention maps are isolated like the "boat" and "mountain" in \Fref{fig:attn} are categorized into separate groups due to small overall. Therefore, we have a total of $N$ groups. Each group has a dedicated branch, resulting in a total of $N>2$ branches. %

\subsubsection{Inversion-Free Multi-Branch Editing}
\label{sec:multi_branch}
We use a set of $N$ branches indexed by $n$. These $N$ branches are in addition to a source branch (also shown in Figure~\ref{fig:pipeline}) that undergoes a DDCM sampling process \citep{xu2023inversion}. The $n^\text{th}$ branch is calibrated to its $(n-1)^\text{th}$ branch, and the first branch is calibrated to the source branch. The $N-$way target branch calibration can occur simultaneously, saving significant compute time. For the DDCM sampling process of the $n^\text{th}$ branch, it has the form of Section~\ref{sec:prelim}, Step~\ref{item:first_step}:

\begin{equation}
\label{eq:multi_sampling}
\overbrace{
    \textcolor{white}{\underbrace{\textcolor{red}{z(n)^\text{edt}}}_{\textcolor{red}{\text{\footnotesize edited latent}}}}
    = \Big(
        \textcolor{white}{\underbrace{\textcolor{blue}{z(n)_\tau^\text{edt}}}_{\textcolor{blue}{\text{\footnotesize noisy latent}}}}
        - \sqrt{1-\alpha_\tau} \big(
            \textcolor{white}{\underbrace{\textcolor{green}{\epsilon(n)_\tau^\text{edt} - \epsilon(n-1)_\tau^\text{edt}}}_{\textcolor{green}{\text{\footnotesize parameterized noise}}}}
            + \textcolor{white}{\underbrace{\textcolor{purple}{\epsilon(n)_\tau^\text{cons}}}_{\textcolor{purple}{\text{\footnotesize consistency noise}}}}
        \big)
    \Big) / \sqrt{\alpha_\tau}
}^{\textcolor{black}{\text{Updating $n^\text{th}$ branch}}}
\end{equation}

Let us break down Eq.~\ref{eq:multi_sampling} step by step. $n=1$ representing the source branch, we have $\textcolor{red}{{z(1)}^\text{edt}} = z^\text{src}$ and $\textcolor{green}{{\epsilon(1)}^\text{edt}_\tau} = {\epsilon}^\text{src}_\tau$. Also, $\textcolor{blue}{z(1)_\tau^\text{edt}}=z_\tau^\text{src}$, which at time step $\tau=\tau_1$, is random noise drawn from $\mathcal{N}(0, \mathbf{I})$. Similarly, when $n=N$, $\textcolor{red}{{z(N)}^\text{edt}}$ represents the final calibrated/edited image containing all the required aspect edits after repeating for $\tau \in \{\tau_1, \tau_2, \hdots \tau_T \}$ timesteps. The noise addition on any target branch remains the same as Step~\ref{item:second_step}, i.e., $\textcolor{blue}{{z(n)}_\tau^\text{edt}} = \sqrt{\alpha_\tau}\textcolor{red}{{z(n)}^\text{edt}} + \sqrt{1-\alpha_\tau}\epsilon$ where $\epsilon\sim\mathcal{N}(0, \mathbf{I})$. For $1<n<N$, we have $\textcolor{green}{{\epsilon(n)}_\tau^\text{edt}} = \epsilon_\theta(\textcolor{blue}{{z(n)}_\tau^\text{edt}}, \tau)$, where $\epsilon_\theta$ represents a parameterized noise predictor network (details in the Appendix \Sref{sec:impl}). A key observation is that the difference in the parameterized noise at the $n^\text{th}$ branch and $(n-1)^\text{th}$ branch is utilized to calculate $\textcolor{red}{{z(n)}^\text{edt}}$ in \eqref{eq:multi_sampling}. %With $N$ branches, these pairwise differences are calculated in a single step.
Finally, $\textcolor{purple}{{\epsilon(n)}_\tau^\text{cons}}$ is defined by $\textcolor{purple}{{\epsilon(n)}_\tau^\text{cons}} = (\textcolor{blue}{{z(n)}_\tau^\text{edt}} - {\sqrt{\alpha_\tau} \textcolor{red}{\hat{z}(n-1)^\text{edt}}})/ \sqrt{1-\alpha_\tau}$. Unlike the dual-branch setup in \cite{xu2023inversion}, the reference initial input is the estimated latent from the previous branch at a previous diffusion denoising iteration as indicated by $\textcolor{red}{\hat{z}(n-1)^\text{edt}}$.

\subsubsection{Cross-Branch Interactions} \label{sec:cross-branch}
%{\bf Rigid Local Branches:} 
For {\bf rigid local branches}, the cross-attention map $\mM^i_n$ from the previous branch is either switched or injected into the current branch, akin to the method used in P2P~\cite{hertz2022prompt}. This approach facilitates local edits while preserving structural consistency. %{\bf Non-Rigid Local Branches:} 
For {\bf non-rigid local branches}, we observe that the query features in the shallow layers of UNet~\cite{ronneberger2015u} can effectively query correlated local contents and textures from the prior branch's latent features, ensuring consistency. Consequently, the key and value features from the prior branch are retained in the current branch to maintain consistent editing. We use a non-rigid editing branch to manage non-rigid local edits. In the current branch $n$, textures from the previous branch $(n-1)$ are preserved by replacing the $K_{n-1}$ and $V_{n-1}$ features from the last branch with the $K_n$ and $V_n$ features in the current branch. Only the query features are preserved to maintain layout semantic correspondence. Additionally, the attention mask $\mM_{n-1}$ from the previous branch's cross-attention layer is used to guide the editing process by adding it to $\mM_{n}$, thereby converting the object layout from $\mM_{n-1}$ to $\mM_{n}$. This step is crucial for object removal or shape modification edits, where the object mask is derived from the previous branch. %{\bf Global Branches:} 
For all {\bf global branches}, there is no replacement of attention features or masks, and the attention mask is not used to guide the editing process, as the entire image is intended to be altered.

\section{Experiments}

%\VISHNU{AT THE END OF EACH SECTION IN EXPERIMENTS, PLEASE ADD THE FOLLOWING TAKEAWAYS RESPECTIVELY}
%\textbf{Takeawy #1: Single Attribute Edit is better than Multiple Attribute Edit
%Takeaway #2: Sequential Multiple Attribute Edit is better than Multiple Attribute Edit
%Takeaway #3: the Parallel edit is better than sequential edit on piebench++ 
%}

% In the section, we first introduce the PIE-Bench++ dataset tailored to enable comprehensive evaluation of multifaceted image editing. Subsequently, we describe the implementation details and evaluation metrics. Then, we compare our proposed method with the state-of-the-art method both quantitatively and qualitatively. 
%To evaluate the performance of multi-aspect image editing, we introduce a new dataset, PIE-Bench++, derived from PIE-Bench ~\cite{ju2023direct} with an extension from single-aspect editing to multi-aspect editing. We then present the evaluation metrics which include the {standard metrics from \cite{ju2023direct} and two newly proposed metrics for aspect accuracy}. We also compare ParallelEdits with several state-of-the-art image editing methods on our PIE-Bench++ dataset. All experiments are conducted on NVIDIA A6000 GPUs. 

\noindent{\bf PIE-Bench++ Dataset}.
We introduce a new dataset, PIE-Bench++, derived from PIE-Bench ~\cite{ju2023direct} and dedicated to evaluate the performance of multi-aspect image editing. 
%Our objective is to address the challenge of multi-aspect image editing. However, the absence of a dedicated benchmark for multi-aspect editing poses a hurdle in evaluating the efficacy of our proposed method. 
The PIE-Bench dataset contains $700$ images and prompts with single-aspect editing including object-level manipulations (addition, deletion, or alteration), attribute-level manipulations (changes in content, pose, color, and material), and image-level manipulations that modify background and overall style. 
{Our PIE-Bench++ extends PIE-Bench by enabling multi-aspect edits: 57\% of our dataset have two aspect edits per prompt, 19\% have more than two edits, and the remaining 24\% have a signle aspect edit. } %most samples in PIE-Bench++ provides a minimum of two aspect edits for each prompt. }A summary of the PIE-Bench++ dataset is presented in \Tref{tab:dataset}. 
{For additional details and examples of the PIE-Bench++ dataset, please refer to the supplementary material.}%For further details and comprehensive examples from the PIE-Bench++ dataset, the reader is directed to consult the supplement.

% \subsection{Implementation details} 

% \begin{figure}[hb]
%     \centering
%     \includegraphics[width=1.0\linewidth]{figs/teaser.pdf}
%   \vspace{-1.5em}
%   \caption{\small \textbf{Existing models fail to handle multi-aspect image editing, while our ParallelEdits enables accurate editing for multiple attributes across different editing categories.} }
%     \label{fig:compare}
%     % \vspace{-0.45cm}
%   \end{figure}

\def\subFigSz{1\linewidth} %width=0.24\linewidth
\begin{figure}[t]
% \vskip -0.2 in
\centering
\includegraphics[width=\subFigSz]{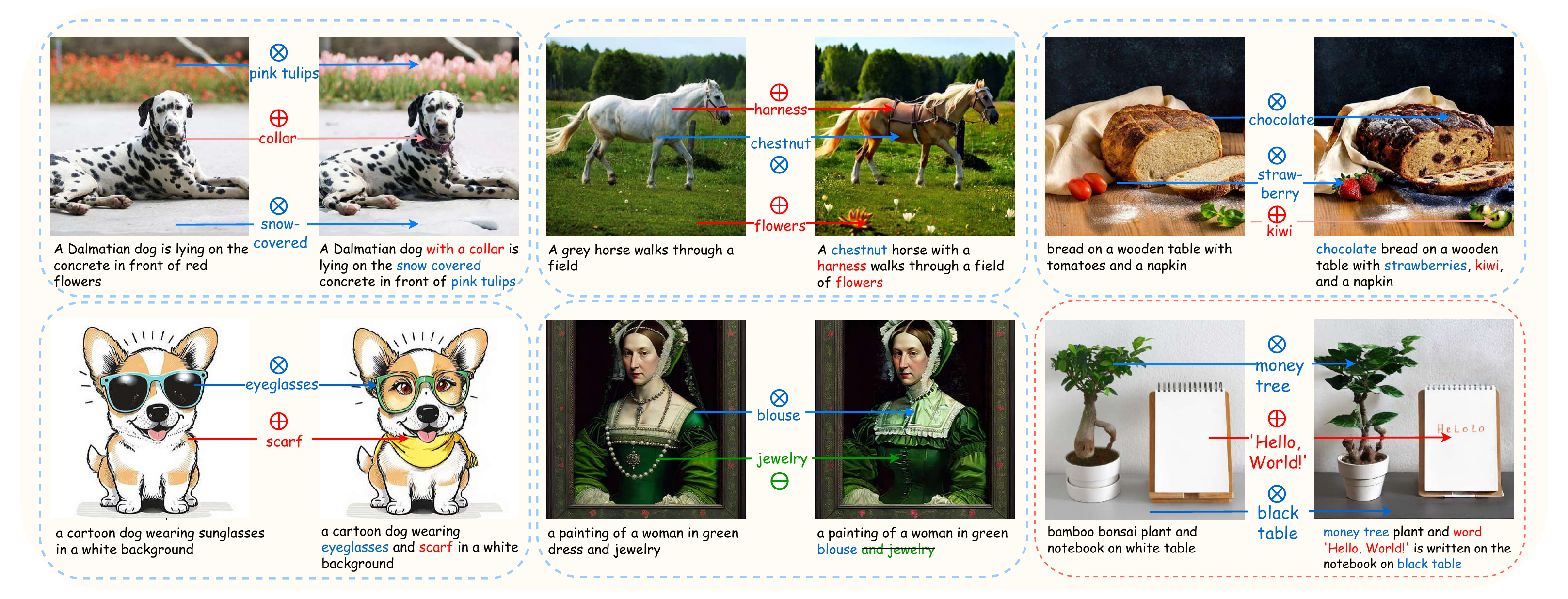} 
% \hskip 0.1in
% \includegraphics[width=\subFigSz]{figs/vis1_jialing.pdf} \hskip 0.1in
% \includegraphics[width=\subFigSz]{figs/vis2.pdf}\\
% \includegraphics[width=\subFigSz]{figs/vis3.pdf} \hskip 0.1in
% \includegraphics[width=\subFigSz]{figs/vis4.pdf}\\
% \includegraphics[width=\subFigSz]{figs/failure_case1.pdf} \hskip 0.1in
% \includegraphics[width=\subFigSz]{figs/failure_case2.pdf}\\
% \vskip 0.1in
% \vskip -0.1 in
\caption{\footnotesize \textbf{Qualitative results of ParallelEdits.} We denote the edits in arrows with edit actions and aspects for each pair of images. The last image pair is a failure case of ParallelEdits.} % 
\vskip -0.1 in
\label{fig:quali}
\end{figure}
\smallskip
\noindent{\bf Evaluation Metrics}. %Evaluating the multi-aspect text-driven image editing method is challenging since most image-language editing models primarily assess the similarity between the entire image and a text prompt. Establishing a method to evaluate the success of each aspect in multi-aspect editing is essential for constructing a comprehensive multi-aspect editing benchmark. In this section, w
We introduce two new metrics designed for evaluating multi-aspect text-driven image editing, alongside standard evaluation metrics.
%\myheading{Standard metrics}. We use the standard metrics for evaluating text-image similarity and image quality. PSNR, LPIPS~\cite{zhang2018unreasonable}, MSE, and SSIM~\cite{wang2004image} are reported to evaluate the image quality. We use CLIP~\cite{CLIP_radford2021} to measure the image-text alignment performance. Additionally, the bi-directional CLIP (D-CLIP) score proposed in \cite{wu2022unifying} has been reported, which is define as:
%\begin{equation}
%\begin{aligned}
%    \mathcal{S}_{\text {D-CLIP }}({\mI_{src}, \mI_{edt}}, \mP_{src}, \mP_{edt})= 
%     \cos \langle &\operatorname{CLIP}_{\text {img }}(\mI_{edt})-\operatorname{CLIP}_{\text {img }}(\mI_{src}), \\
%     &\operatorname{CLIP}_{\text {text }}(\mP_{edt})-\operatorname{CLIP}_{\text {text}} (\mP_{src}) \rangle 
%\end{aligned}
%\end{equation}

%\subsection{Aspect editing accuracy}
%\mingzhen{need double-check}

\myheading{(a) Aspect Accuracy-LLaVA}. Drawing inspiration from the remarkable capability of large vision language models in comprehending intricate semantics within images, we propose to innovatively leverage them as an ``omniscient'' agent equipped with extensive knowledge to understand various attributes of images. We use the LLaVA~\cite{liu2023llava} model, trained on visual grounding tasks, to evaluate the accuracy of multi-aspect image editing. Given a text prompt with multiple aspects, such as ``\textit{A [pink] [taxi] with [colorful] [flowers] on top}'', we provide the following prompt with the edited image to the LLaVA model: ``\textit{Does the image match the elements in [ ]: A [pink] [taxi] with [colorful] [flowers] on top? Return a list of numbers where 1 is matched and 0 is unmatched.}'' We then parse the returned list and compute its average to determine the aspect accuracy. We name this new evaluation metric as \textit{AspAcc-LLaVA}. Examples and detailed explanations of this evaluation metric are available in the supplementary material. 

\myheading{(b) Aspect Accuracy-CLIP}. We also use the similarity of the CLIP~\cite{CLIP_radford2021} to evaluate if an attribute has been successfully edited. %Recall the notation from \Sref{sec:method}, 
Given an edited image {$\mathcal{I}_{edt}$} and the target prompt $\mathcal{P}_{edt}$ with {$k$} edited aspects $\mA_{edt}$, every time we remove an aspect $\mA^j_{edt}$ from $\mathcal{P}_{edt}$ and revert it back to $\mA^i_{src}$ as $\hat{\mP}_{edt}$. We then extract the CLIP~\cite{CLIP_radford2021} similarity between the edited image $I_{edt}$ and two prompts, i.e., $s_1 = CLIP(\mathcal{I}_{edt}, \mathcal{P}_{edt})$ and $s_2 = CLIP(\mathcal{I}_{edt},  \hat{\mP}_{edt})$. We expect $s_1 > s_2$ if the aspect $\mA^j_{edt}$ has been successfully edited. {Thus, the aspect accuracy is $\frac{k_s}{k}$ when a total of $k_s$ aspects have been successfully edited among $k$ target edits.} Note that in the case of an edited or added object that also involves changes in attributes (such as color or material), we consider it a successful edit only if both the object and its attributes have been successfully modified. We name this metric as \textit{AspAcc-CLIP}.

\myheading{(c) Standard Metrics}. Several standard metrics widely used for evaluating text-image similarity and image quality are considered, including PSNR, LPIPS~\cite{zhang2018unreasonable}, MSE, and SSIM~\cite{wang2004image}. We also use the CLIP~\cite{CLIP_radford2021} score to measure the image-text alignment performance. Additionally, the bi-directional CLIP (D-CLIP) score \cite{wu2022unifying} is reported, which is formulated as follows:
\[\small 
     \cos \langle \operatorname{CLIP}_{\text {img }}(\mI_{edt})-\operatorname{CLIP}_{\text {img }}(\mI_{src}), \operatorname{CLIP}_{\text {text }}(\mP_{edt})-\operatorname{CLIP}_{\text {text}} (\mP_{src}) \rangle
\]
% which is a number between $[0,1]$.

\subsection{Quantitative Results}
We first conduct experiment on the PIE-Bench++ dataset to compare our method with the state-of-the-art text-driven image editing methods combining their corresponding inversion method leads to best performance, including DDIM+MasaCtrl~\cite{cao_2023_masactrl}, DDIM+Prompt-to-Prompt (P2P)~\cite{hertz2022prompt}, DDIM+Plug-and-Play (PnP)~\cite{tumanyan2023plug}, StyleDiffusion (StyleD)~\cite{wang2023stylediffusion}+P2P,
Null-text Inversion (NTI)~\cite{mokady2023null}+P2P, DirectInverison (DI)\cite{ju2023direct}+PnP, and InfEdit \cite{xu2023inversion}. {An intuitive way to improve off-the-shelf image editing methods is to apply the single-aspect editing method sequentially. We follow ~\cite{joseph2024iterative} to adapt existing image editing methods into sequential editing processes, where these methods are applied multiple times to achieve multi-aspect editing. Each time, only one aspect is edited. } \Tref{tab:main} presents the metrics in terms of text-image similarity (i.e., CILP and D-CLIP scores), computational efficiency, and aspect accuracy.
Our ParallelEdits model outperforms all baselines in editing effectiveness, with a slightly longer runtime than the InfEdit model. {Even though sequential editing better aligns the target prompt than their vanilla methods, it significantly increases computational overhead and may propagate editing errors over time. Moreover, although the sequential editing is conducted in the latent space, it would introduce more noise and artifacts to the edited image. Hence, their performance in all editing quality metrics was inferior to our method.} %the preservation of background and image quality in sequential editing are worse compared to \Tref{tab:bg}. }%\textit{Takeaway: sequential editing is not efficient for multi-aspect editing due to noise, artificial and error accumulating.} 

\begin{figure}[t]
    \centering
    \includegraphics[width=1.0\linewidth]{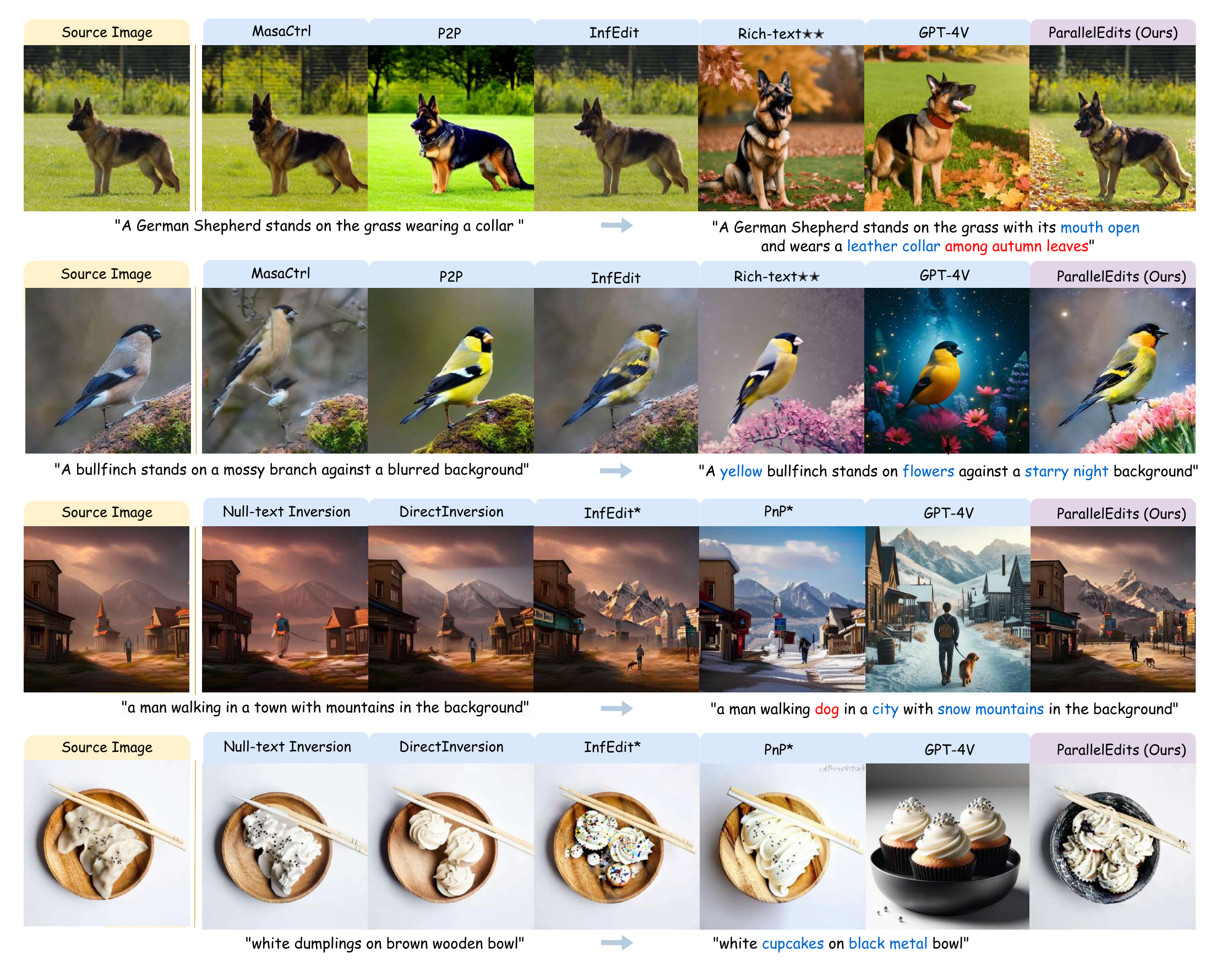}
   \vspace{-2em}
    \caption{\footnotesize \textbf{Qualitative results comparison.}  
  Current methods fail to edit multiple aspects effectively, even using sequential edits (noted as *). Methods marked with \(\star\)\(\star\) taking additional inputs other than source image and plain text.}
    \label{fig:comapre}
    \vspace{-0.3cm}
  \end{figure}  
\begin{table*}[t]
	\centering
	\footnotesize
	\setlength\tabcolsep{2pt}
	\renewcommand\arraystretch{1.0}
       \scalebox{0.91}{
	\begin{tabular}{l|ccccccc|cccc|c}
		% \rowcolor{mygray}
         & \textbf{StyleD} & \textbf{MasaCtrl} & \textbf{P2P} & \textbf{DI} & \textbf{NTI} & \textbf{InfEdit} & \textbf{PnP}  & \textbf{DI*} & \textbf{P2P*} & \textbf{InfEdit*} & \textbf{PnP*} & \textbf{Ours} \\ \hline \hline
        % \rowcolor{mygray}
         {CLIP (\%)   $\uparrow$} & 24.02 & 23.37 & 24.00 & 24.40 & 24.03 & 24.44 & {24.90} & 22.80 & 25.13 & \underline{25.17} & 25.39 & \textbf{25.70} \\
         {D-CLIP (\%)  $\uparrow$} & 8.43 & 7.68 & 11.43 & \underline{13.23} & 12.08 & 11.02 & 11.83 & 2.74 & 8.30 & 11.77 & 11.85& \textbf{20.70} \\
         {Eff. (secs/sample) $\downarrow$} & 382.98 & 12.70 & 33.72 & 29.70 & 145.29 & \textbf{2.22} & 32.51 & 100.98 & 121.32 & 11.82 & 122.81 & \underline{4.98} \\
         {AspAcc-CLIP (\%) $\uparrow$} & 32.37 & 34.05 & 26.14 & 31.95 & 42.19 & 42.38 & {44.91} & 28.23 & 38.96 & 42.38 & \underline{48.20} & \textbf{51.05} \\
         {AspAcc-LLaVA (\%)$\uparrow$} & 53.79 & 55.79 & 55.04 & 54.42 & 59.80 & 60.55 & {61.36} & 46.24 & 55.21 & 61.90 & \underline{63.80} & \textbf{65.19} \\ \hline \hline
	\end{tabular}}
    % \vspace{-0.09in}
    \caption{\footnotesize \textbf{Comparison results in multi-aspect image editing on the PIE-Bench++ dataset.} Computational efficiency is abbreviated as Eff., and * denotes the method using sequential editing. The best performance is highlighted in {\textbf{bold}} and the second best performance is {\underline{underlined}}.}
	\vspace{-0.3cm}
	\label{tab:main}
\end{table*}
  
\subsection{Qualitative Results}
\vspace{-0.2cm}
{Fig. \ref{fig:quali} presents several examples of our method's multi-aspect editing on the PIE-Bench++ dataset. The results demonstrate the effectiveness of our method in handling multiple and varied types of edits across diverse image content. Fig. \ref{fig:comapre} further compares our method with several state-of-the-art models and one popular multi-modal large language model, GPT-4V~\cite{Gpt-4v}, by providing the source image, source prompt, and target prompt to guide the image editing. The Rich-text~\cite{ge2023expressive} model differs from other models, which uses rich-text prompt to edit the image generated from the plain (source) text prompt. The results show that current image editing models even with sequential editing fail to edit multiple aspects, while multi-modal large language models fail to preserve the content of source image. Our method achieves visually convincing results by successfully editing different attributes with good content preservation.}

\vspace{-0.1cm}
\subsection{Ablation Study and Analysis}
\vspace{-0.2cm}

\begin{figure}[t]
    % \vskip -0.2 in
    \centering
    \includegraphics[width=1.0\linewidth]{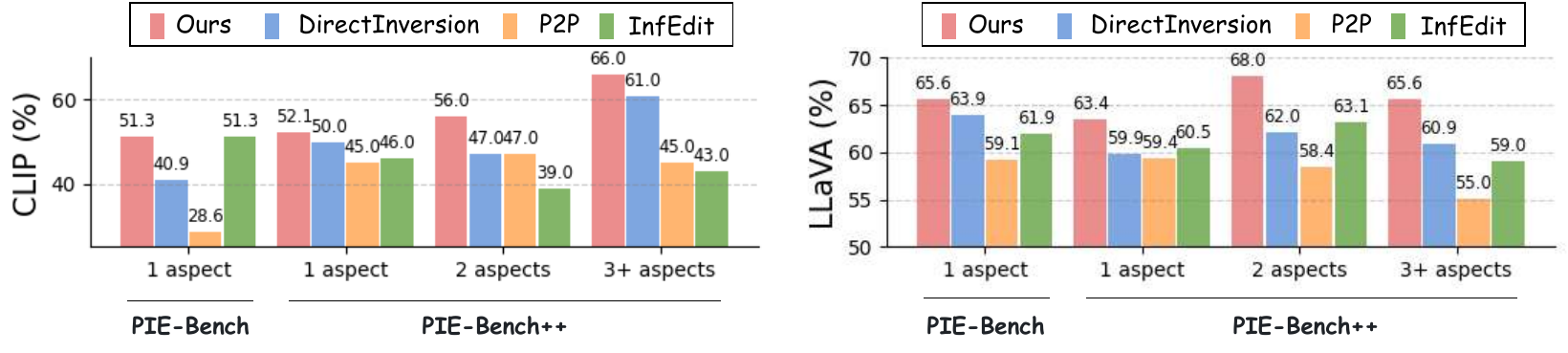} \hskip 0.1in
    % \vskip 0.1in
    % \vskip -0.1 in
    \vspace{-1em}
    \caption{\footnotesize \textbf{Comparison across different numbers of editing aspects.} We also include the comparison in PIE-Bench dataset. Our proposed method is robust to different numbers of editing aspects.} % 
    \label{fig:aspect}
    % \vspace{-0.15in}
\end{figure}

\myheading{(a) Impact of Editing Aspect Number. }{We first examine the performance of our ParallelEdits and baseline methods on various editing aspect numbers by comparing CLIP and LLaVA-based aspect accuracies on the original PIE-Bench~\cite{ju2023direct} and our PIE-Bench++ datasets. The bar charts in Fig. \ref{fig:aspect} show the outstanding performance of our method across all settings, including single-aspect editing on two datasets and multi-aspect editing. %We also compare with other methods in PIE-Bench dataset~\cite{ju2023direct} which contains only single aspect to demonstrate that the performance of our method does not downgrade in editing single aspect.
\textit{Takeaway: the proposed ParallelEdits demonstrates robustness across varying numbers of editing aspects.}}

\myheading{(b) Evaluation on Perservation. }%Ability across different method}.
We follow \cite{ju2023direct} to evaluate the background preservation. We first use the PSNR, LPIPS~\cite{zhang2018unreasonable}, MSE and SSIM~\cite{wang2004image} to evaluate the background preservation. We measure that metric on a subset of images of our proposed PIE-Bench++ dataset where the background can be well defined in that image, e.g., no image style or background editing, and the background is visible after aspect editing. The results are shown in \Tref{tab:bg}, where we compare our method with the top performance methods in \Tref{tab:main}. Moreover, we adopt the similar way as calculating the AspAcc-LLaVA to prompt LLaVA~\cite{liu2023llava} for evaluating how the unchanged aspect preserves in the edited image. We also calculate the CLIP~\cite{CLIP_radford2021} score between the target image and the text prompt after removing all edited aspects. The results are reported in \Tref{tab:bg} noted as CLIP and LLaVA, respectively. \textit{Takeaway: preservation is even maintained in ParallelEdits}. 

\begin{table*}[t]
	\centering
	\small
	% \resizebox{0.49\textwidth}{!}{
	\setlength\tabcolsep{3pt}
	\renewcommand\arraystretch{1.0} 
	\begin{tabular}{l|cccc||cc}
		% \hline \thickhline 
		% \rowcolor{mygray}
         % & Structure  & Similarity & \multicolumn{2}{c}{Aspect Accuracy}   \\
        %  \rowcolor{mygray}
        &\multicolumn{4}{c}{\textbf{Background Preservation}}& \multicolumn{2}{c}{\textbf{Aspect Preservation\%}}\\
		% \rowcolor{mygray}
         Methods    & PSNR$\uparrow$ & LPIPS$_{\times10^3}$$\downarrow$&MSE$_{\times10^4}\downarrow$ & SSIM$_{\times10^2}\uparrow$ &CLIP$\uparrow$ & LLaVA $\uparrow$\\ \hline \hline
		
        % MasaCtrl & 22.993 & 95.766 &  &\\
        P2P~\cite{hertz2022prompt} & 18.48 / \textcolor{green}{16.64} & 188.26 / \textcolor{green}{231.83} & 190.07 / \textcolor{green}{345.07} &73.55 / \textcolor{green}{69.17} & 20.72 / \textcolor{green}{23.48} & 66.59 / \textcolor{green}{72.60} \\
         % & $\checkmark$ & 16.64 & 231.83 & 345.07 &69.17 &23.48 &72.60\\
        % Null-text Inversion &  27.708  & 53.186 &   27.7213 &  85.69\\
        % DirectInversion & 24.885  &72.79  & 51.324 & 83.610 \\
        PnP~\cite{tumanyan2023plug}&22.73 / \textcolor{green}{21.54} & 103.16 / \textcolor{green}{120.87} & {\textbf{75.97}} / \textcolor{green}{102.47} & 80.73 / \textcolor{green}{78.85} &20.79 / \textcolor{green}{\textbf{25.59}} &75.65 / \textcolor{green}{78.77}\\
        % &$ \checkmark$ & 21.54 & 120.87 & 102.47 & 78.85 &{\textbf{25.59}} &78.77 \\
        InfEdit~\cite{xu2023inversion}&  24.61 / \textcolor{green}{24.09}& 103.99 / \textcolor{green}{107.43} & 160.54 / \textcolor{green}{163.72} & 78.85 / \textcolor{green}{79.64} &24.69 / \textcolor{green}{25.04} &75.90 / \textcolor{green}{78.05}\\
        % & $\checkmark$&  24.09 & 107.43 & 163.72 & 79.64 &25.04 &78.05\\ \hline
		Ours & {\textbf{26.13}} & {\textbf{95.87}} & 113.86 & {\textbf{82.35}} & 25.49 & {\textbf{80.70}}\\ \hline \hline
		\end{tabular}
	% }
	\captionsetup{font=small}
    \vspace{-0.01in}
    \caption{\footnotesize \textbf{Comparison results in terms of background and aspects preservation.} The results from sequential editing is noted as \textcolor{green}{green}. ParallelEdits achieves state-of-the-art performance on multi-aspect editing while preserving the background and content consistency. }%P}
    \vspace{-0.1in}
	\label{tab:bg}
 % \vspace{-in}
\end{table*}
% The best performance is highlight in \textcolor{blue}{blue} and the second best performance in \textcolor{green}{green}.

%     \begin{figure}[t]
%     % \vskip -0.2 in
%     \centering
%     \includegraphics[width=1.0\linewidth]{figs/failure.pdf} \hskip 0.1in
%     % \vskip 0.1in
%     % \vskip -0.1 in
%     \vspace{-2em}
%     \caption{\footnotesize \textbf{Failure cases}. ParallelEdits fails to edit text and dramatic background.} % 
%     \label{fig:fail}
%     \vspace{0.1in}
% \end{figure}
 \vspace{0.05in}

\myheading{(c) Branches numbers and aspect grouping.} To demonstrate the effectiveness of our multi-branch design and early aspect grouping, we design additional ablation studies for our method in threefold. (1) We only use one single non-rigid branch to conduct all edits; (2) we remove the aspect categorization process from the pipeline and use the same non-rigid branch for each edit; (3) we adopt one single branch for different type of edits without using any auxillary branches which results a total of three branches (also see Section~\ref{sec:aux} for more details). \textit{Takeaway: As shown in \Tref{tab:grouping}, the multi-branch design and aspect grouping play a significant role in enhancing the performance of our proposed ParallelEdits}. 

\begin{table*}[t]
	\centering
	\small
	\resizebox{0.9\textwidth}{!}{
	\setlength\tabcolsep{3pt}
	\renewcommand\arraystretch{1.0}
	\begin{tabular}{lccc|cccc}
		% \hline \thickhline
		% \rowcolor{mygray}
        &with aspect&with aspect & with auxillary& \multicolumn{2}{c}{\textbf{Similarity \%} } & \multicolumn{2}{c}{\textbf{Aspect Accuracy \%}}   \\
         % \rowcolor{mygray}
         &categorization&grouping & branch & CLIP$\uparrow$& D-CLIP$\uparrow$ &   CLIP$\uparrow$ & LLaVA $\uparrow$\\ \hline \hline
       \multirow{4}{*}{ParallelEdits} & $\times$
        & $\times$& $\times$   &24.32&10.45& 40.97 &  57.67 \\ 
       & $\times$&$\checkmark$&$\checkmark$&25.14&11.97&46.66 & 58.37\\
        &  $\checkmark$ & $\times$&$\times$  & 24.50  & 12.33&48.08 &61.22 \\
        
        &  $\checkmark$&$\checkmark$&$\checkmark$&   {\textbf{25.70}} &{\textbf{20.70}}  & {\textbf{51.05}}  & 	{\textbf{65.19}}  
        % w/o grouping \\
        \\ \hline \hline
		\end{tabular}
	}
	\caption{\footnotesize \textbf{Ablation studies on branch numbers and aspect grouping.}}
 % \vspace{-0.8cm}
	\label{tab:grouping}
%  \vspace{-0.3in}
\end{table*}

 \vspace{-0.2in}
\myheading{(d) Performance comparison on each category.} Recall that our dataset includes nine different categories for editing. We compare the performance of baseline models and our approach across the nine categories, as presented in \Tref{tab:cat}. \textit{Takeaway: Our proposed ParallelEdits achieves state-of-the-art performance across most categories.}

\begin{table}[t]
    \centering
    \label{tab:data}
    \setlength\tabcolsep{3pt}
    \scalebox{0.96}{ 
    \begin{tabular}{l|ccccccc|c|c} % Note the 'r' for right-aligned third column 

    % \rowcolor{mygray}
     & \multicolumn{7}{c|}{\textbf{Change}} & \textbf{Add} & \textbf{Delete} \\
    % \rowcolor{mygray} 
     \textbf{Asepct Acc-CLIP} & Object & Content & Pose & Color & Material & Background & Style& Object & Object \\ \hline\hline
    P2P~\cite{hertz2022prompt} & 33.13 & 20.00 & 25.83 & 34.17 & 31.67 & 30.63 & 19.38 & 22.29 & 11.88 \\
    MasaCtrl~\cite{cao_2023_masactrl} & 40.83 & 23.75 & \textbf{40.83} & 20.00 &30.83 & 26.88 &29.38 & 37.08 & 28.96 \\
    NTI~\cite{mokady2022null} & \underline{48.13} & \underline{41.25} & 23.75 & \underline{51.25} & 24.17 & 51.25 &22.50 & 40.42 & 32.08 \\
    DirectInversion~\cite{ju2023direct} & 40.63 & 26.25 & 23.33 & 40.00 & 25.42 & 32.50 &25.00 & 30.00 & 20.83 \\
    InfEdit~\cite{xu2023inversion} & 36.24  & 33.33 & 25.41 & 41.67 & 27.50 &48.75 & 41.88& \underline{50.63} & \underline{45.41} \\
    PnP~\cite{tumanyan2023plug}  & 44.38 & 27.29 & 27.91 & 49.17 & \underline{32.91} & \underline{52.50} &\textbf{55.63} &44.38 & 42.08 \\ \hline
    ParallelEdits  & \textbf{51.46} & \textbf{44.16} & \underline{39.58} & \textbf{60.00} & \textbf{47.50} & \textbf{60.00} &\underline{50.00} & \textbf{56.04} & \textbf{52.08} \\ \hline \hline

    \end{tabular}}
    \vspace{0.06in}
    \caption{\footnotesize \textbf{Comparison on each category in PIE-Bench++.} Our ParallelEdits achieves the best performance on most of the categories from the dataset.}
    \label{tab:cat}
    \end{table}

\myheading{Limitations and Failure Cases.}
The proposed ParallelEdits has several limitations. First, it cannot handle the text editing in the image, as shown in the last image pair of \Fref{fig:quali}. % there is distortion for the edited text. 
Second, ParallelEdits fails to edit dramatic background changes, as {examples shown in the supplementary material}.%shown in the last images of \Fref{fig:vis}).
% \VISHNU{Please add the following limitations a) Failure cases are possible.}

% \vspace{-0.2in}
\section{Conclusion} 
 % \vspace{-0.1in}

In this work, we propose a new research task, multi-aspect text-driven image editing, to modify multiple object types, attributes, and relationships. 
We introduce a dedicated method, ParallelEdits, to multi-aspect text-driven image editing as an effective and efficient solution to this problem. 
Due to the lack of evaluation benchmark, we introduce PIE-Bench++, an improved version of PIE-Bench~\cite{ju2023direct} tailored for simultaneous multiple-aspect edits within images. ParallelEdits achieves better quality and performance than existing methods on proposed PIE-Bench++. 
Our work introduces ParallelEdits, a novel approach that adeptly handles multiple attribute edits simultaneously, preserving the quality of edits across single and multiple attributes through a unique attention grouping mechanism without adding computational complexity. 
There are several future works we would like to explore. First, different aspects of an image have a specific semantic order. Editing these aspects according to their intrinsic order will simplify the editing process. Secondly, the current ParallelEdits still has limitations, as shown in \Fref{fig:quali}. It will be of interest to study approaches to improve these aspects. 

\myheading{Ethics Statement}. In anticipation of contributing to the academic community, we plan to make the dataset and associated code publicly available for research. Nonetheless, we acknowledge the potential for misuse, particularly by those aiming to generate misinformation using our methodology. We will release our code under an open-source license with explicit stipulations to mitigate this risk. These conditions will prohibit the distribution of harmful, offensive, or dehumanizing content or negatively representing individuals, their environments, cultures, religions, and so forth through the use of our model weights.
% Our aim is to foster responsible use that advances research while safeguarding against unethical applications. 

\myheading{Acknowledgement}. This work was supported in part by the National Science Foundation (NSF) Projects under grants SaTC-2153112, No.1822190, and TIP-2137871. Prof. Lokhande thanks support provided by University at Buffalo Startup funds. We thank Sudhir Kumar Yarram for the insightful discussions on the project.

% ---- Bibliography ----
%
% BibTeX users should specify bibliography style 'splncs04'.
% References will then be sorted and formatted in the correct style.
%
\clearpage
\bibliographystyle{unsrt}

\bibliography{pubs}

\clearpage
\appendix

\section*{Appendix}
\section{ParallelEdits: The Algorithm}
In this section we provide Algorithm~\ref{alg:group}: \textit{Early Aspect Grouping} and Algorithm~\ref{alg:cap}: \textit{ParallelEdits on a particular branch}. These algorithms describe the overall idea behind ParallelEdits. They are also pictorially illustrated in Figures $2$ and $3$ of the main paper. Let us denote an arbitrary branch and the timestep in the diffusion process by $n$ and $t$ respectively. Firstly, in Algorithm~\ref{alg:group}, we demonstrate how \textit{Early Aspect Grouping} is conducted over the attention maps. Recall that we refer to this as ``early" aspect grouping because only a few steps (maximum of $5$) are sufficient to perform the grouping. This phase of ParallelEdits takes as an input, the edit action set $\{E^{i\rightarrow j}\}$ and the corresponding cross-attention maps for every token $\A_{src}^j$, and outputs the grouped edit actions set $\bar{\mA}^{c}_{edt}$. Recall from Section 4 of the paper that $E^{i \rightarrow j} \in \{\textcolor{blue}{\boldsymbol{\otimes}}, \textcolor{red}{\boldsymbol{\oplus}}, \textcolor{green}{\boldsymbol{\ominus}}, \textcolor{black}{\boldsymbol{\oslash}} \}$, with $\textcolor{blue}{\boldsymbol{\otimes}}$ denoting a swap action, $\textcolor{red}{\boldsymbol{\oplus}}$ denoting an add action, $\textcolor{green}{\boldsymbol{\ominus}}$ denoting aspect deletion, and $\textcolor{black}{\boldsymbol{\oslash}}$ indicating no change in the aspect. Once grouped edit actions set is computed, it is fed into Algorithm~\ref{alg:group} to conduct multi-aspect editing and obtain the edited latent features. 
In Algorithm 2, we implement several operations on the attention masks, similar to the P2P method~\cite{hertz2022prompt}, and describe them as follows.
%In Algorithm~\ref{alg:cap}, we make use of a few action on the attention masks, similiar to P2P~\cite{hertz2022prompt}, and define them below:

\myheading{Replace}: Swapping token attention mask $\mM_{n-1}$ in the prompt from previous branch, overriding $\mM_{n}$;

\myheading{Refine}: Injecting only the attention mask that corresponds to the unchanged part of the prompt from $\mM_{n-1}$ to $\mM_{n}$;

\myheading{Retain}: Keeping the attention mask $\mM_{n}$  unchanged.

%Both Algorithm~\ref{alg:group} and Algorithm~\ref{alg:cap} have been described in the main paper, corresponding to Figures 2 and 3. 

% \mingzhen{define branch nums, branch n prev branch n-1, }
% \end{algorithm*}

% \SetKwComment{Comment}{/* }{ */}

\renewcommand{\algorithmicrequire}{\textbf{Input:}}
\renewcommand{\algorithmicensure}{\textbf{Output:}}
%\vspace{-0.2in}
\begin{algorithm}[!ht]
    \begin{minipage}{\linewidth}
        \caption{Early Aspect Grouping}\label{alg:group} 
        \begin{algorithmic}[1]
            \Require Edit action set $\{E^{i \rightarrow j}\}$, {Cross attention maps $\{{\mM}\}$}
            
            \State rigid-edit $\gets \{\}$, non-rigid-edit $\gets \{\}$, global-edit $\gets \{\}$
            % \Set
            % \State $\bar{\mA}^{c}_{edt} \gets \{\}$
            \For{${\mA}^{i \rightarrow j}_{edt} \in \{E^{i \rightarrow j}\}$ }
                \If{$\gamma(\bar{\mM}_{edt}^j) \geq \beta\gamma(\sum\{\bar{\mM}_{edt}\})$}  \Comment{This is a global edit}
                \State global-edit $\gets$ global-edit + $\{{E}^{i \rightarrow j}\}$
                
                \ElsIf{$\phi(\bar{\mM}_{src}^i, \bar{\mM}_{edt}^j) < \theta$}  \Comment{This is a rigid edit}
                    \For{$\bar{\mA}^{c}_{edt} \in$ rigid-edit}
                    \If {mIoU($\bar{\mA}^{c}_{edt}, E^{i \rightarrow j} \geq \theta$)} \Comment{$\bar{\mA}^{c}_{edt}$ is a set of grouped edit actions}
                        \State $\bar{\mA}^{c}_{edt} \gets \bar{\mA}^{c}_{edt} + E^{i \rightarrow j}$
                    \Else
                        \State rigid-edit $\gets$ rigid-edit + $E^{i \rightarrow j}$
                    \EndIf
                    \EndFor
                % \State rigid-edit $\gets$ rigid-edit + ${\mA}^{i \rightarrow j}_{edt}$
                \ElsIf {$\phi(\bar{\mM}_{src}^i, \bar{\mM}_{edt}^j) \geq \theta$}
                \Comment{This is a non-rigid edit}
                     \For{$\bar{\mA}^{c}_{edt} \in$ non-rigid-edit}
                        \If {mIoU($\bar{\mA}^{c}_{edt}, E^{i \rightarrow j} \geq \theta$)}
                            \State $\bar{\mA}^{c}_{edt} \gets \bar{\mA}^{c}_{edt} + E^{i \rightarrow j}$
                        \Else
                            \State non-rigid-edit $\gets$ non-rigid-edit + $E^{i \rightarrow j}$
                        \EndIf
                    \EndFor
                \EndIf
            \EndFor

            \Ensure Grouped edit actions set $\{\bar{\mA}^c_{edt}\}$
        \end{algorithmic} 
    \end{minipage}
\end{algorithm}
% \vspace{-0.2in}

\begin{algorithm}[t]
    \begin{minipage}{\linewidth}
        \caption{ParallelEdits on a Particular Branch }\label{alg:cap} 
        \begin{algorithmic}[1]
            \Require {Denoising UNet $\varepsilon_\theta$,
            \renewcommand{\algorithmicrequire}{\hskip\algorithmicindent}
            \Require Grouped edit action $\bar{\mA}^{c}_{edt}$,  \Comment{Output from early aspect grouping}
            \Require Latent feature in previous branch and previous timestep $z_{n-1}^t, z_{n}^{t-1}$,
            \Require Cross attention maps $\{{\mM}\}$, 
            \Require Self attention features $Q_{n-1}, K_{n-1}, V_{n-1}$,
            \Require Edit type list: rigid-edit, non-rigid-edit, global-edit}
            % \State \textbf{Input:}
            % \textcolor{blue}{Grouped edit action $\bar{\mA}^{c}_{edt}$, }
            % \State \hskip\algorithmicindent \textcolor{blue}{Denoising UNet $\varepsilon_\theta$}
            % \State \hskip\algorithmicindent \textcolor{blue}{Diffusion timestep $t$}
            % \State \hskip\algorithmicindent  \textcolor{blue}{Latent feature in previous branch and previous timestep $z_{n-1}^t, z_{n}^{t-1}$}
            % \State \hskip\algorithmicindent  \textcolor{blue}{Cross attention maps $\{{\mM}\}$}
            % \State \hskip\algorithmicindent  \textcolor{blue}{Self attention features $Q_{n-1}, K_{n-1}, V_{n-1}$ }
            % \State \hskip\algorithmicindent  \textcolor{blue}{Edit type list: 
            % rigid-edit, non-rigid-edit, global-edit}
            \State $\mM_n \gets \varepsilon_\theta(\bar{\mA}^{c}_{edt}, z_{n}^{t-1}, t-1)$
            \If{$\bar{\mA}^{c}_{edt} \in$ global-edit} \Comment{This is a global edit}
                \State retain($\mM_n$) \Comment{Do not switch attention maps for global edits}
            \ElsIf{$\bar{\mA}^{c}_{edt} \in$ non-rigid-edit}  \Comment{This is a non-rigid edit}
                \State replace($\mM_{n-1}, \mM_{n}$ )
            \ElsIf{$\bar{\mA}^{c}_{edt} \in$ rigid-edit} \Comment{This is a rigid edit}
                \State $\{Q_n,K_n,V_n\} \gets \{Q_n,K_{n-1},V_{n-1}\}$
                \State refine($\mM_{n-1}, \mM_{n}$ )
            
            \EndIf
        \State $\bar{\mM}_n \gets \text{binarize}(\sum_{m=0}^{m \leq n}\mM_m)$
        \State $z_n^t \gets \bar{\mM}_n \odot z^t_{n} + (1-\bar{\mM}_n)\odot z^t_{n-1}$
        \Ensure  Latent feature $z_n^t$
        \end{algorithmic} 
    \end{minipage}
\end{algorithm}
 % \vspace{in}

\section{Some More Details on ParallelEdits}

%\subsection{Latent Consistency Model}% and Inversion-Free}
In the literature \cite{cao_2023_masactrl, hertz2022prompt}, image editing processes have been conducted through the implementation of a dual-branch approach. This method involves utilizing a source and target branches for editing. Specifically, the source branch is reverted to $z_0$, while the trajectory of the target branch is iteratively adjusted. By computing the distance from the source branch and $\epsilon^{\text{cons}}$ with Latent Consistency Model~\cite{song2023consistency}, the target branch is calibrated at each time step.
\\
Our experiments, as seen in Section~$5$ of the main paper, show the ineffectiveness of a dual-branch procedure for multi-aspect editing tasks. Specifically, a single target branch is inadequate, leading to imperfection in the target image. Thereby we advocate multi-aspect editing through the use of multiple target branches. Each target branch handles a group of aspects, with simpler aspects such as non-rigid local edits directed to initial branches, and more complex aspects such as rigid local edits deferred to subsequent ones. Note that however, all the branches operate simultaneously.  \\

%In the following section, we will delve deeper into this concept.

% PLEASE ADD EQUATIONS 7-12 from ALGORITHM 2 of INFEDIT OVER HERE. ALTERNATIVELY, INSTEAD OF EQUATIONS HAVING IT AS FIGURE 2 of \url{https://arxiv.org/pdf/2402.13490v1.pdf} IS ALSO OKAY

\myheading{Auxiliary Rigid / Non-Rigid Branches.}
\label{sec:aux}
In the main paper, it was noted that there was one dedicated branch for each type of edit: non-rigid, rigid, and global edit. The Early Aspect Grouping algorithm~\ref{alg:group} classifies aspects into these three categories. Our experiments revealed that sometimes, due to low overlap between attention maps, aspects may not always be grouped into dedicated rigid or non-rigid branches. In such cases, it becomes necessary to include an auxiliary branch to handle the ungrouped aspects. Therefore, ParallelEdits may involve a single rigid branch and additional auxiliary branches to manage ungrouped aspects, and similarly, a single non-rigid branch and supplementary auxiliary branches to address ungrouped aspects. An ablation study on auxiliary branches is provided in \Tref{tab:grouping}.
%Recall that in the main paper, the proposed ParallelEdits method categorizes all editing actions into different edit types (i.e., non-rigid, rigid, and global edit), and further groups the aspects to combine those that can be edited together within the same branch. However, there are instances when grouping does not yield successful results, for instance, when two edits lack significant overlap. In such cases, we duplicate this branch to handle those edits independently. %However, sometimes the grouping is not successful, e.g., two edits do not share significant overlapping, in this case, we duplicate this branch to separately handle those edits. 
%We demonstrate the effectiveness of this design in \Tref{tab:grouping} empirically. The experimental results show that the multi-branch pipeline improves the multi-aspect editing capability of the model. 
% {\color{Please write a take-away of table1 here}}. 

\section{More Details on Evaluation Metrics}
In this section, we describe more details of our evaluation metrics.

\begin{figure}[!hb]
    \centering
    \includegraphics[width=1.0\linewidth]{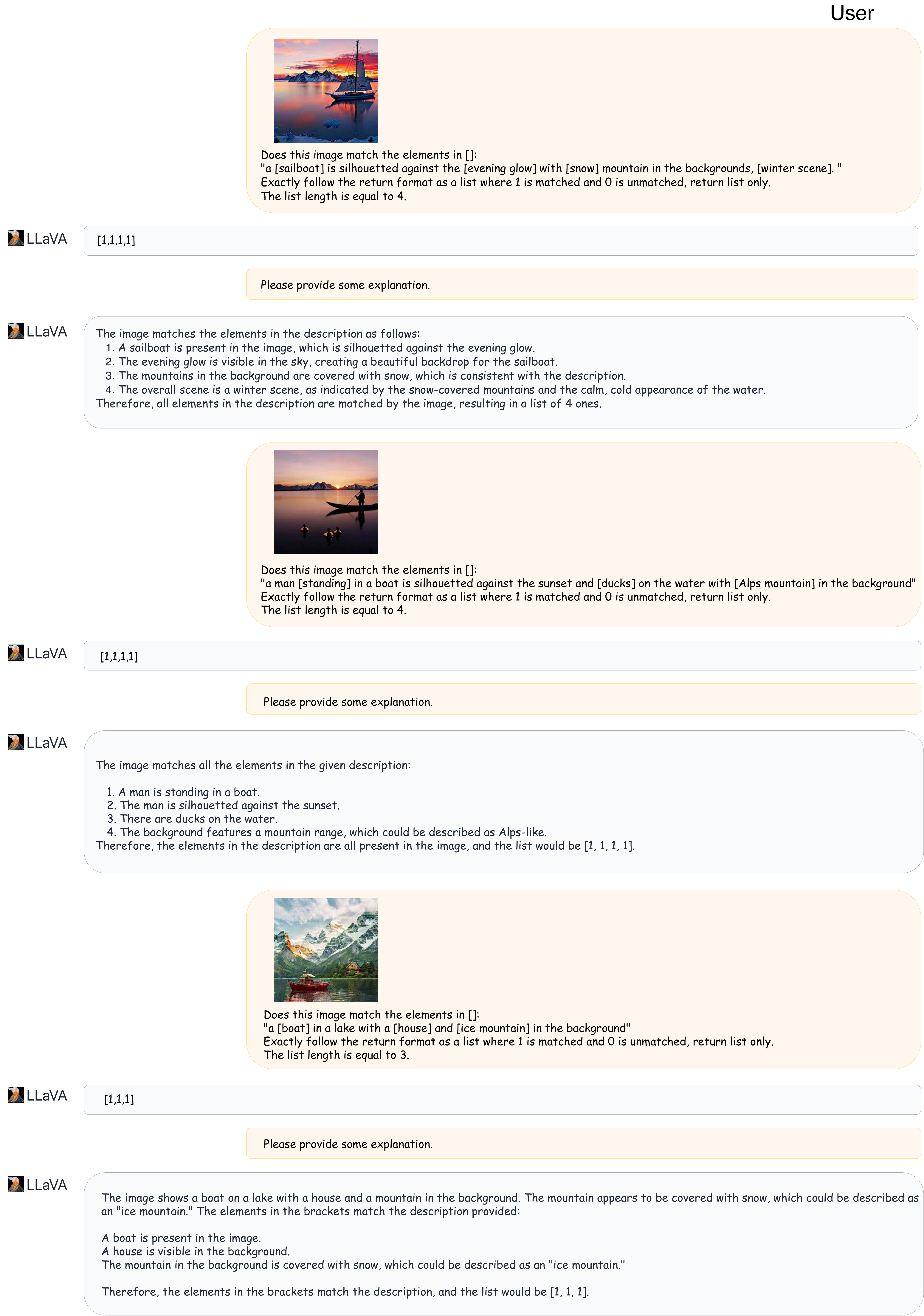}
  % \vspace{-1.5em}
    \caption{\footnotesize \textbf{Examples of prompting LLaVA for aspect accuracy measurement in cases of successful editing.} LLaVA can effectively illustrate if and how the target image and edits are misaligned.}
    \label{fig:llava}
    \vspace{-0.45cm} 
  \end{figure}

\begin{figure}[!ht]
    \centering
    \includegraphics[width=1.0\linewidth]{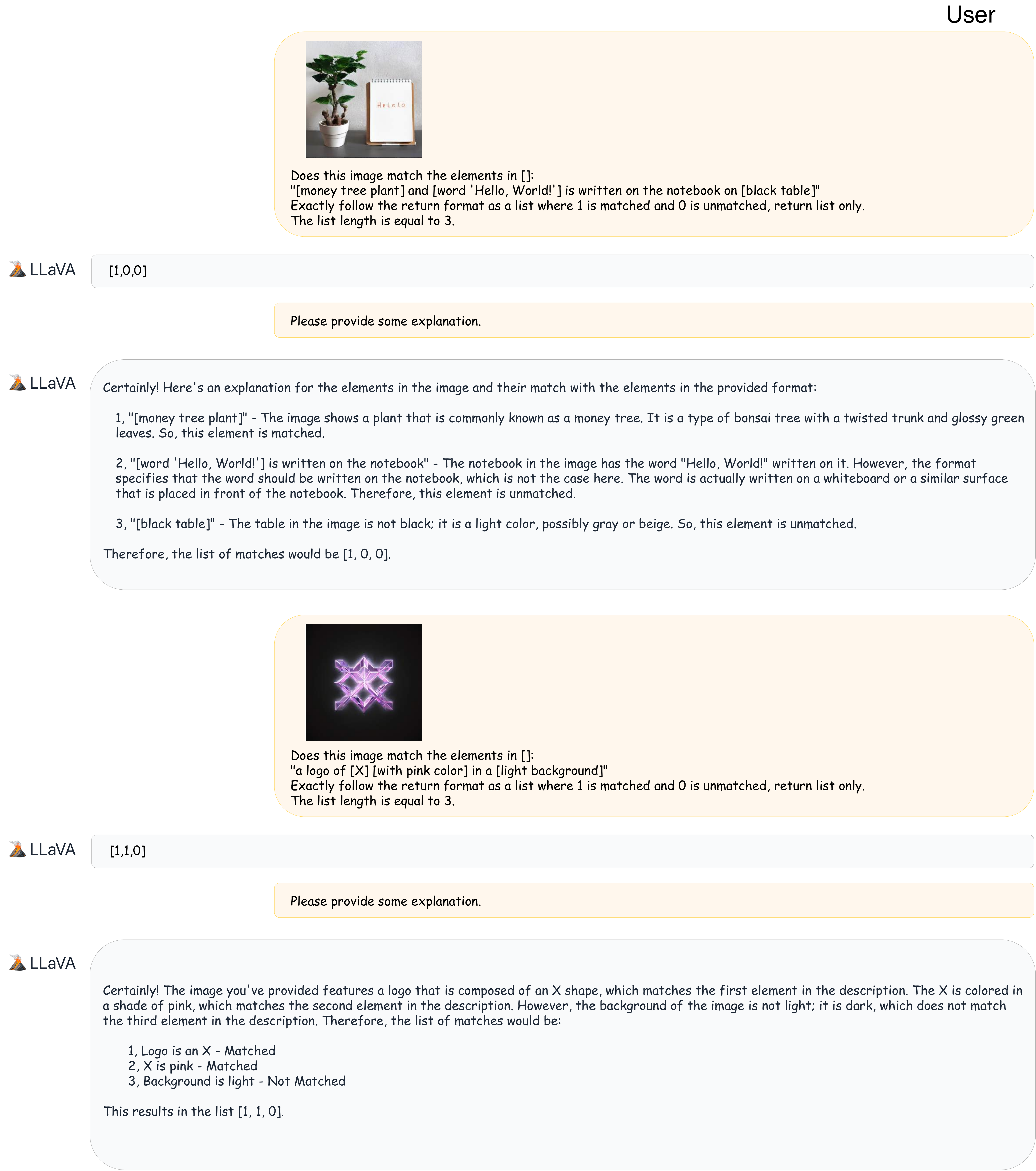}
  % \vspace{-1.5em}
    \caption{\footnotesize \textbf{Examples of prompting LLaVA for aspect accuracy measurement in cases of unsuccessful editing.}}
    \label{fig:llava_fail}
    \vspace{-0.45cm}
  \end{figure}

\myheading{LLaVA aspect accuracy}.
We show how we leverage LLaVA~\cite{liu2023llava} to evaluate the multi-aspect editing accuracy in~\Fref{fig:llava} and \Fref{fig:llava_fail}, we also prompt LLaVA~\cite{liu2023llava} for explanation to human readers. LLaVA~\cite{liu2023llava} could provide detailed summary for the image and also explanations for the mismatched between edited aspects and image.

% \vspace{-1.5em}

\myheading{Other evaluation metrics}.
Moreover, even though the Structure Distance~\cite{tumanyan2022splicing} has been used in PIE-Bench~\cite{ju2023direct} to evaluate the structure between source and target image while ignoring appearance information, it could not serve as a good evaluation metric for multi-aspect editing. This is due to the fact that the structure of multi-aspect edited target image may necessitate substantial modifications, particularly when it involves adding or removing multiple objects.

\section{Implementation Details}
\label{sec:impl}
Our proposed ParallelEdits is based on the Latent Consistency Model~\cite{song2023consistency}, with the publicly available LCM \footnote{\href{https://huggingface.co/SimianLuo/LCM_Dreamshaper_v7}{https://huggingface.co/SimianLuo/LCMDreamshaperv7}} which is finetuned from Stable Diffusion v1.5. We then follow \cite{xu2023inversion} to leverage their proposed inversion-free technique in ParallelEdits for image editing. During sampling, we perform LCM sampling~\cite{song2023consistency} with 15 denoising steps, and the classifier-free guidance (CFG) is set to 4.0. ParallelEdits can control the editing strength by adjusting the CFG . There's a trade-off between achieving satisfactory inversion and robust editing ability. A higher CFG tends to produce stronger editing effects but may lower inversion results and identity preservation.
We also set the hyper-parameter $\theta$ as 0.9 and $\beta$ as 0.8 in our experiments, where $\theta, \beta$ are used to determine the edit type of a given edit action.

In the inversion-free multi-branch editing approach, for $1<n<N$, the noise estimation is also conditioned on a text conditioning $c_n$ in branch $n$. This can be expressed as $\textcolor{green}{{\epsilon(n)}_\tau^\text{edt}} = \epsilon_\theta(\textcolor{blue}{{z(n)}_\tau^\text{edt}}, \tau, c_n)$. Here, $c_1$ corresponds to the source prompt, $c_N$ corresponds to the target prompt, and $c_n$ represents the prompt that includes all aspect edits up to branch $n$.

%where $c_n$ is adopted all edit from branch $1$ to branch $n$, so we have$\textcolor{green}{{\epsilon(n)}_\tau^\text{edt}} = \epsilon_\theta(\textcolor{blue}{{z(n)}_\tau^\text{edt}}, \tau, c_n)$
% Moreover, 
% \myheading{Inversion techniques}.

%\clearpage
\section{Additional Details of PIE-Bench++}
% \mingzhen{Need revised}

\subsection{PIE-Bench++ Details} 
Unlike existing benchmarks that primarily focus on single-aspect edits, PIE-Bench++ is tailored to multiple aspect edits, reflecting the complexities inherent in real-world editing tasks. Our enhanced dataset, PIE-Bench++, builds upon the PIE-Bench~\cite{ju2023direct} by incorporating 700 images across nine diverse categories, covering both natural and artificial scenes, with a significant focus on multi-aspect editing scenarios. Specifically, the Change Object category involves swapping objects in the scene with different yet reasonable alternatives. Add Object adds new elements to the scene. Delete Object focuses on removing objects, testing the model's ability to erase elements seamlessly. Change Object Content alters the content of specific objects, such as changing the design on a shirt or the pattern on a wall. Change Object Pose includes changes in the shape of objects, humans, or animals. Change Object Color assesses the model's ability to apply accurate color changes. Change Object Material evaluates the rendering of different textures and materials. Change Background involves editing scenarios where there is a distinct foreground object and a main background. This type of edit focuses on seamlessly integrating new background elements while preserving the integrity of the foreground object. Change Image Style involves the application of style transfer techniques to the entire image while ensuring the original content remains intact. For example, this could involve transforming a photograph to adopt a cartoon style. Each category is carefully curated to provide a comprehensive evaluation of the dataset's multi-aspect editing capabilities, the summary of the dataset is shown in \Tref{tab:dataset}.

\subsection{Dataset Annotation}
The annotation process involves a primary annotator who labels the source prompt, describing the original image, and the target prompt, which outlines the desired modifications to generate the target image. The target prompt is carefully annotated to include all editing pairs expected to be reflected in the target image. Subsequently, a second annotator reviews the annotations for accuracy and consistency, ensuring the reliability of the dataset.
The majority of target prompts in PIE-Bench++ feature at least two edited aspects. Nevertheless, within the categories that solely changing background and image styles, the number of edits is usually constrained to one or two aspects. This limitation is due to the intrinsic characteristics of these attributes, such as each image having only one background or style.

\begin{figure}[ht]
    \centering
    \includegraphics[width=1.0\linewidth]{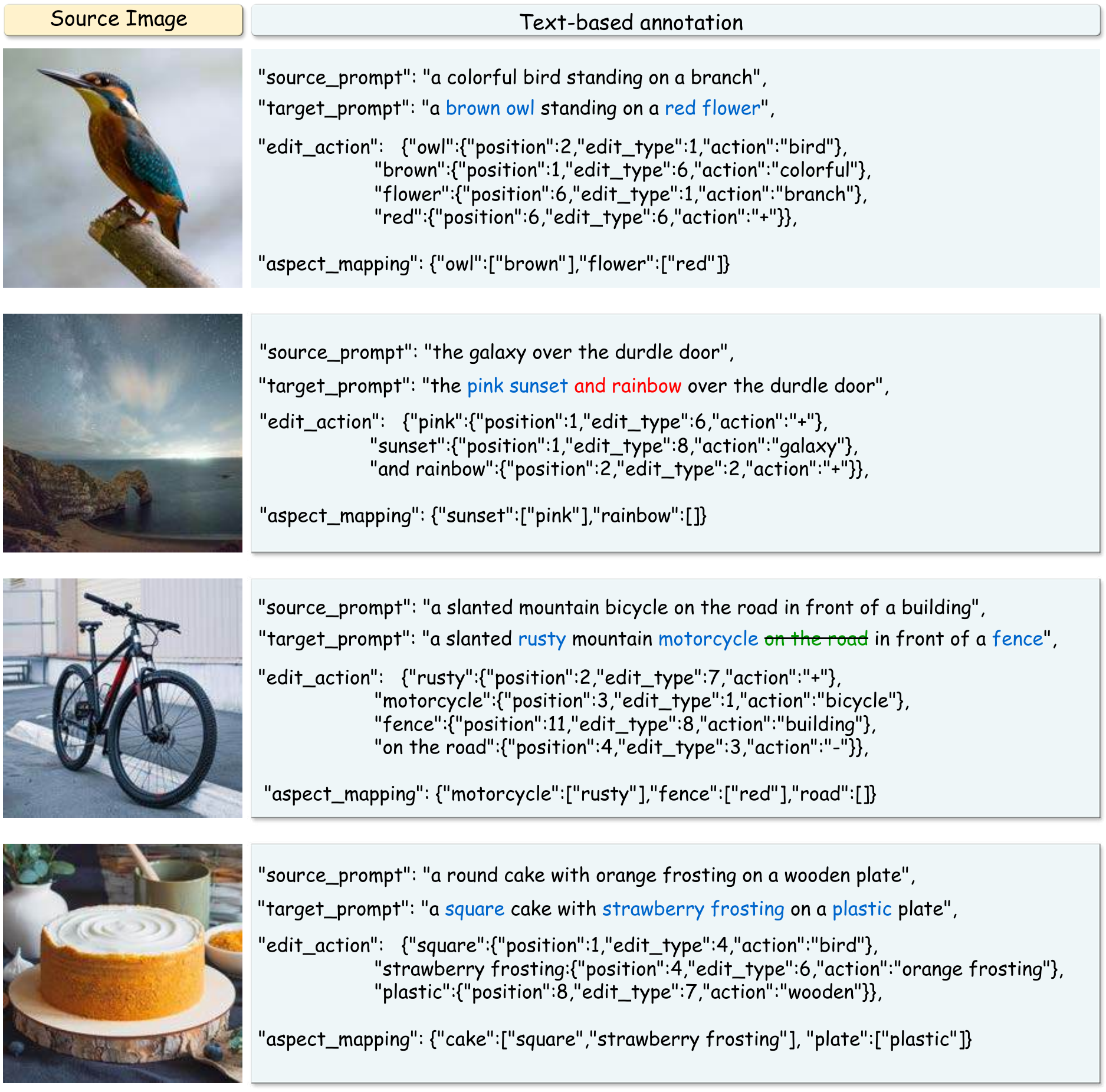}
  % \vspace{-1.5em}
    \caption{\footnotesize \textbf{Annotation examples from PIE-Bench++}. 
    Each annotation containing a Source Prompt, Target Prompt, Edit Action, and Aspect Mapping. Edit action contains the specific instructions including the desired modification index in source prompt as position, edit type among 9 catergories and the action $\in \{\textcolor{blue}{\boldsymbol{\otimes}}, \textcolor{red}{\boldsymbol{\oplus}}, \textcolor{green}{\boldsymbol{\ominus}}\}$. The aspect mapping indicts the pair between object and attribute. }
    \label{fig:anno}
    \vspace{-0.45cm}
  \end{figure}

\myheading{Annotation format details}.
Each image in the dataset annotation is associated with key elements as shown in \Fref{fig:anno}: a source prompt, a target prompt, an edit action, and a mapping of aspects. The edit action specifies the position index in the source prompt where changes are to be made, the type of edit to be applied, and the operation required to achieve the desired outcome. The aspect mapping connects objects undergoing editing to their respective modified attributes, enabling the identification of which objects are subject to editing.

   \begin{table}[t]
    \centering
    \label{tab:data}
    \setlength\tabcolsep{3pt}
    \scalebox{0.9}{ 
    \begin{tabular}{r|ccccccc|c|c} % Note the 'r' for right-aligned third column 

    % \rowcolor{mygray}
     & \multicolumn{7}{c|}{\textbf{Change}} & \textbf{Add} & \textbf{Delete} \\
    % \rowcolor{mygray} 
     & Object & Content & Pose & Color & Material & Background & Style& Object & Object \\ \hline\hline

    \#Edited Aspect & 302 & 98 & 120 & 188 & 99 & 112 &165 & 178 & 119 \\
    \#Edited Token & 316 & 155 & 227 & 205 & 116 & 175 &424 &507 & 381 \\ \hline\hline

    \end{tabular}}\vspace{0.2cm}
    \caption{\footnotesize \textbf{Summary of Editing Types and Categories in PIE-Bench++ dataset.} There are 10 different categories in PIE-Bench++ and a total number of 700 images.}\vspace{-0.4cm}
    \label{tab:dataset}
    \end{table}
    
% Overall, the dataset structure facilitates systematic and comprehensive multi-aspect editing tasks, enabling researchers and practitioners to evaluate and advance editing algorithms effectively.

% This example illustrates the annotation for a data sample within the PIE-Bench++ dataset. Here, the source prompt describes a scene with a man sitting in a boat against a sunset backdrop. The target prompt introduces additional elements such as a standing man, ducks on the water, the Alps mountain, and the sun. The editing instructions specify the changes to be made to the source prompt, including replacing the action 'sitting' with 'standing' for the man, and adding 'ducks on the water', 'Alps', and 'and sun' at the specified positions. Additionally, the obj-adj-pair indicates the corresponding modifications for objects and their associated adjectives.

\clearpage
\section{Additional Qualitative Results}
\begin{figure}[h]
    \centering
    \includegraphics[width=1.0\linewidth]{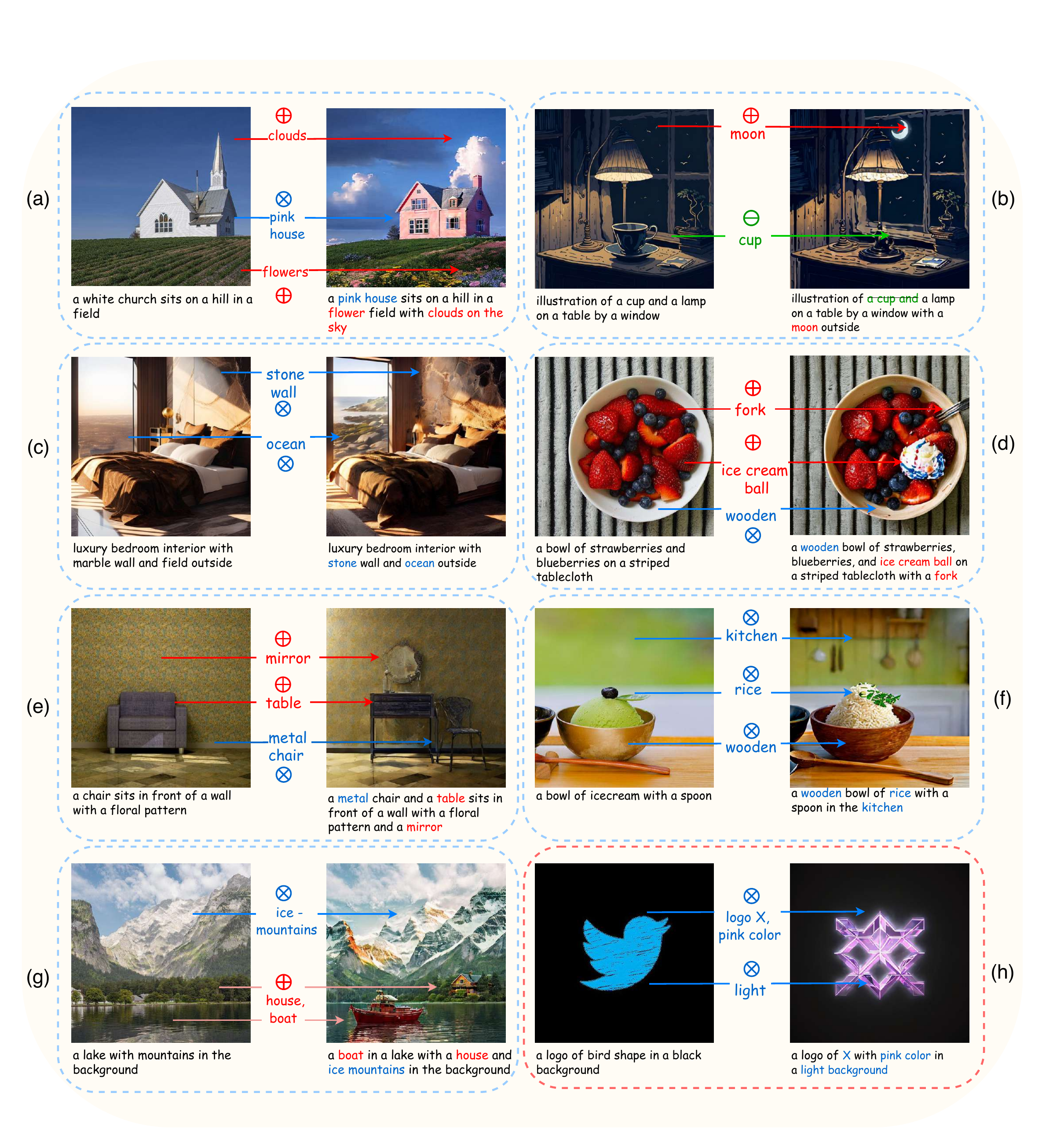}
  % \vspace{-1.5em}
    \caption{\small\textbf{Qualitative results from ParallelEdits.} ParallelEdits is able to swap, add and delete multiple aspects. The last image pair is a failure case of ParallelEdits.}
    \label{fig:vis}
    \vspace{-0.45cm}
  \end{figure}

\begin{figure}[!h]
    \centering
    \includegraphics[width=1.0\linewidth]{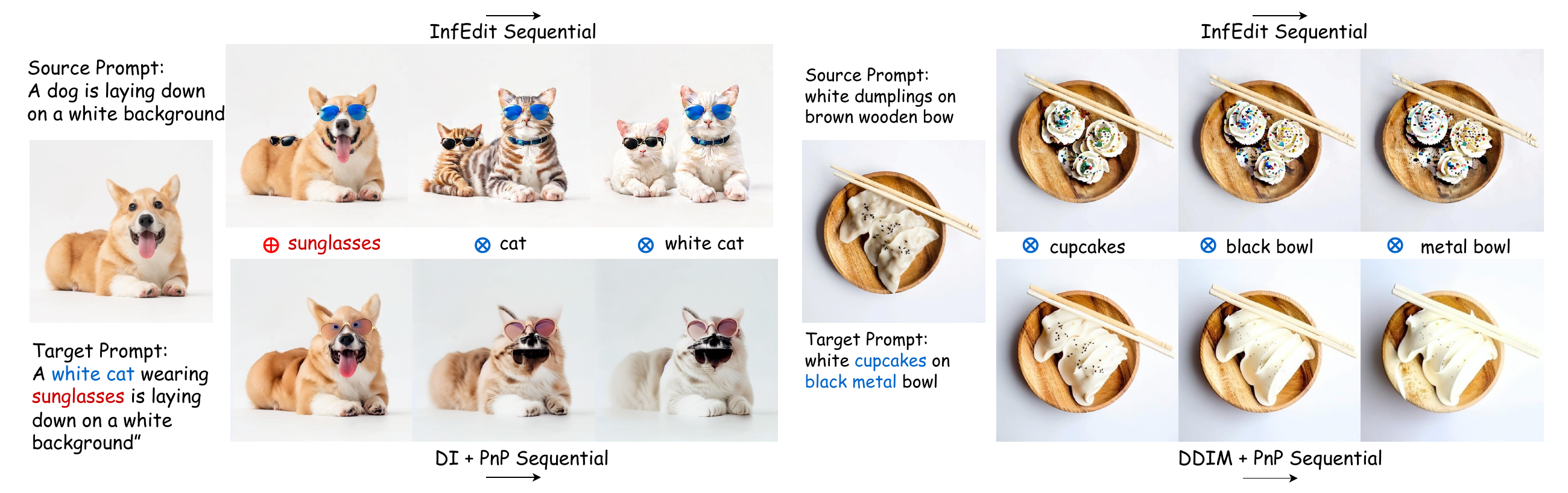}
  \vspace{-2em}
    \caption{\small \textbf{Sequential editing using single-aspect text-driven image editing methods.} The sequential editing might accumulate errors and undo previous edits. It also fails to edit significantly overlapped objects.}
    \label{fig:single_fail}
    \vspace{-0.5cm}
  \end{figure}

\begin{figure}[!h]
    \centering
    \includegraphics[width=0.9\linewidth]{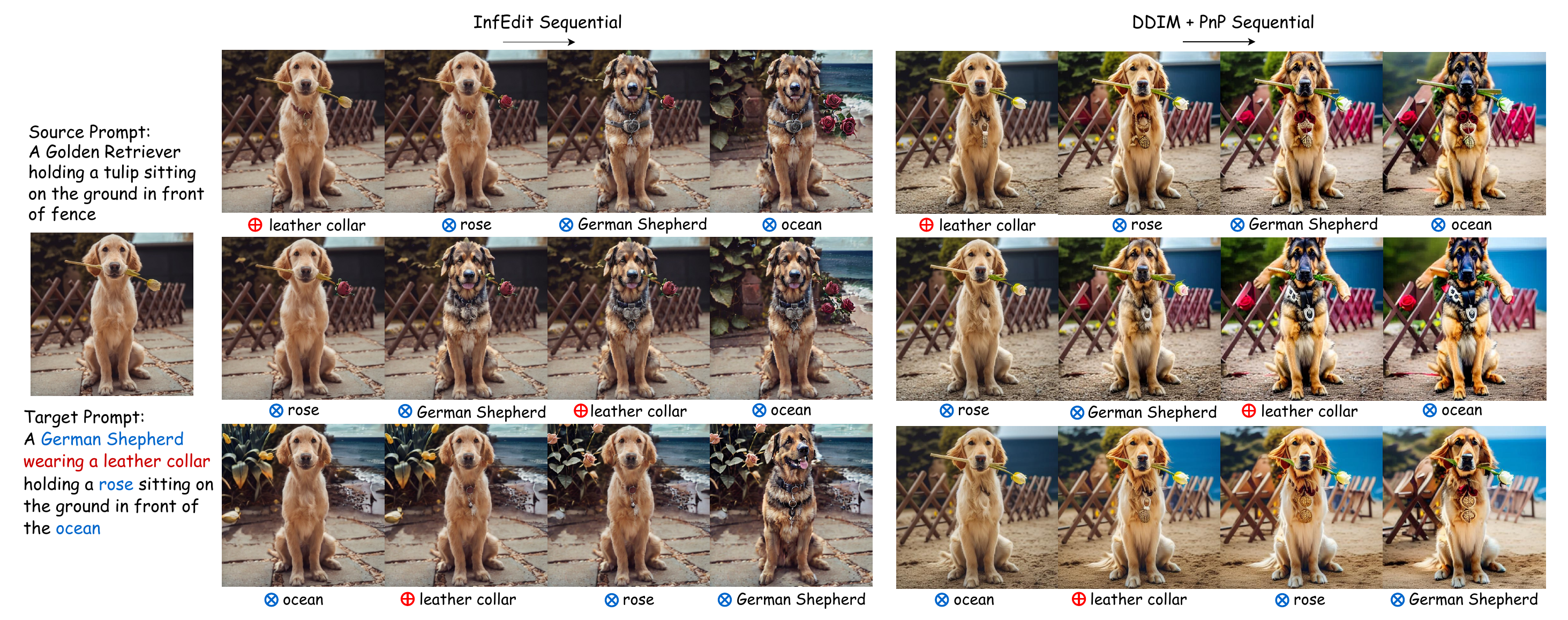}
  % \vspace{-1.5em}
    \caption{\small \textbf{Sequential editing with different orders.} Sequential editing with different orders can yield varying final results. Additionally, it may lead to error accumulation and potentially overwrite previous edits.}
    \label{fig:sequentialo}
    \vspace{-0.45cm}
  \end{figure}

We also provide more qualitative results in \Fref{fig:vis}, showing the effectiveness of our proposed method in handling multi-aspect editing tasks. These examples showcase the model's proficiency in executing intricate edits. For instance, as depicted in \Fref{fig:vis} (b), our method successfully removes a cup while accurately reconstructing the obscured parts of the lamp behind it. In \Fref{fig:vis}  (a), the model demonstrates its ability to swap and add aspects, while preserving the composition of the scene. The results underscore the model's adeptness in interpreting and executing sophisticated editing instructions, leading to visually consistent and contextually fitting edited images. Additional, we also provide the results for sequential editing methods with different editing order in \Fref{fig:single_fail} and \Fref{fig:sequentialo}.

\clearpage
\section*{NeurIPS Paper Checklist}

\begin{enumerate}

\item {\bf Claims}
    \item[] Question: Do the main claims made in the abstract and introduction accurately reflect the paper's contributions and scope?
    \item[] Answer: \answerYes{} % Replace by \answerYes{}, \answerNo{}, or \answerNA{}.
    \item[] Justification: The main claims made in the abstract and introduction accurately reflect the paper's contributions and scope.

\item {\bf Limitations}
    \item[] Question: Does the paper discuss the limitations of the work performed by the authors?
    \item[] Answer: \answerYes{} % Replace by \answerYes{}, \answerNo{}, or \answerNA{}.
    \item[] Justification: We include the limitation and failure cases of the work in Sec. 5.

\item {\bf Theory Assumptions and Proofs}
    \item[] Question: For each theoretical result, does the paper provide the full set of assumptions and a complete (and correct) proof?
    \item[] Answer: \answerNA{} % Replace by \answerYes{}, \answerNo{}, or \answerNA{}.
    \item[] Justification: This paper does not include theoretical result.

    \item {\bf Experimental Result Reproducibility}
    \item[] Question: Does the paper fully disclose all the information needed to reproduce the main experimental results of the paper to the extent that it affects the main claims and/or conclusions of the paper (regardless of whether the code and data are provided or not)?
    \item[] Answer: \answerYes{} % Replace by \answerYes{}, \answerNo{}, or \answerNA{}.
    \item[] Justification: We fully disclose all the information needed to reproduce the main experimental results

\item {\bf Open access to data and code}
    \item[] Question: Does the paper provide open access to the data and code, with sufficient instructions to faithfully reproduce the main experimental results, as described in supplemental material?
    \item[] Answer: \answerYes{} % Replace by \answerYes{}, \answerNo{}, or \answerNA{}.
    \item[] Justification: The code and data will be open-sourced for academic use.

\item {\bf Experimental Setting/Details}
    \item[] Question: Does the paper specify all the training and test details (e.g., data splits, hyperparameters, how they were chosen, type of optimizer, etc.) necessary to understand the results?
    \item[] Answer: \answerYes{} % Replace by \answerYes{}, \answerNo{}, or \answerNA{}.
    \item[] Justification: We specify all the training and test details

\item {\bf Experiment Statistical Significance}
    \item[] Question: Does the paper report error bars suitably and correctly defined or other appropriate information about the statistical significance of the experiments?
    \item[] Answer: \answerYes{} % Replace by \answerYes{}, \answerNo{}, or \answerNA{}.
    \item[] Justification: The paper reports error bars

\item {\bf Experiments Compute Resources}
    \item[] Question: For each experiment, does the paper provide sufficient information on the computer resources (type of compute workers, memory, time of execution) needed to reproduce the experiments?
    \item[] Answer: \answerYes{} % Replace by \answerYes{}, \answerNo{}, or \answerNA{}.
    \item[] Justification: The paper provide sufficient information on the computer resources
    \item[] Guidelines:
    \begin{itemize}
        \item The answer NA means that the paper does not include experiments.
        \item The paper should indicate the type of compute workers CPU or GPU, internal group, or cloud provider, including relevant memory and storage.
        \item The paper should provide the amount of compute required for each of the individual experimental runs as well as estimate the total compute. 
        \item The paper should disclose whether the full research project required more compute than the experiments reported in the paper (e.g., preliminary or failed experiments that didn't make it into the paper). 
    \end{itemize}
    
\item {\bf Code Of Ethics}
    \item[] Question: Does the research conducted in the paper conform, in every respect, with the NeurIPS Code of Ethics \url{https://neurips.cc/public/EthicsGuidelines}?
    \item[] Answer: \answerYes{} % Replace by \answerYes{}, \answerNo{}, or \answerNA{}.
    \item[] Justification: The code follows the NeurIPS Code of Ethics.

\item {\bf Broader Impacts}
    \item[] Question: Does the paper discuss both potential positive societal impacts and negative societal impacts of the work performed?
    \item[] Answer: \answerYes{} % Replace by \answerYes{}, \answerNo{}, or \answerNA{}.
    \item[] Justification: The paper includes the discussion of potential societal impacts.
    
\item {\bf Safeguards}
    \item[] Question: Does the paper describe safeguards that have been put in place for responsible release of data or models that have a high risk for misuse (e.g., pretrained language models, image generators, or scraped datasets)?
    \item[] Answer: \answerNA{} % Replace by \answerYes{}, \answerNo{}, or \answerNA{}.
    \item[] Justification: The paper poses no such risks.

\item {\bf Licenses for existing assets}
    \item[] Question: Are the creators or original owners of assets (e.g., code, data, models), used in the paper, properly credited and are the license and terms of use explicitly mentioned and properly respected?
    \item[] Answer: \answerYes{} % Replace by \answerYes{}, \answerNo{}, or \answerNA{}.
    \item[] Justification: CC-BY 4.0 for PIE-Bench.

\item {\bf New Assets}
    \item[] Question: Are new assets introduced in the paper well documented and is the documentation provided alongside the assets?
    \item[] Answer: \answerYes{} % Replace by \answerYes{}, \answerNo{}, or \answerNA{}.
    \item[] Justification: The documentation provided alongside the assets

\item {\bf Crowdsourcing and Research with Human Subjects}
    \item[] Question: For crowdsourcing experiments and research with human subjects, does the paper include the full text of instructions given to participants and screenshots, if applicable, as well as details about compensation (if any)? 
    \item[] Answer: \answerNA{} % Replace by \answerYes{}, \answerNo{}, or \answerNA{}.
    \item[] Justification: The paper does not involve crowdsourcing nor research with human subjects

\item {\bf Institutional Review Board (IRB) Approvals or Equivalent for Research with Human Subjects}
    \item[] Question: Does the paper describe potential risks incurred by study participants, whether such risks were disclosed to the subjects, and whether Institutional Review Board (IRB) approvals (or an equivalent approval/review based on the requirements of your country or institution) were obtained?
    \item[] Answer: \answerNA{} % Replace by \answerYes{}, \answerNo{}, or \answerNA{}.
    \item[] Justification: The paper does not involve crowdsourcing nor research with human subjects.

\end{enumerate}

\end{document}

%% file: definitions.tex
\def\mA{\mathcal{A}}
\def\mB{\mathcal{B}}
\def\mC{\mathcal{C}}
\def\mD{\mathcal{D}}
\def\mE{\mathcal{E}}
\def\mF{\mathcal{F}}
\def\mG{\mathcal{G}}
\def\mH{\mathcal{H}}
\def\mI{\mathcal{I}}
\def\mJ{\mathcal{J}}
\def\mK{\mathcal{K}}
\def\mL{\mathcal{L}}
\def\mM{\mathcal{M}}
\def\mN{\mathcal{N}}
\def\mO{\mathcal{O}}
\def\mP{\mathcal{P}}
\def\mQ{\mathcal{Q}}
\def\mR{\mathcal{R}}
\def\mS{\mathcal{S}}
\def\mT{\mathcal{T}}
\def\mU{\mathcal{U}}
\def\mV{\mathcal{V}}
\def\mW{\mathcal{W}}
\def\mX{\mathcal{X}}
\def\mY{\mathcal{Y}}
\def\mZ{\mathcal{Z}} 

\def\bbN{\mathbb{N}} 
\def\bbR{\mathbb{R}} 
\def\bbP{\mathbb{P}} 
\def\bbQ{\mathbb{Q}} 
\def\bbE{\mathbb{E}}

\def\1n{\mathbf{1}_n}
\def\0{\mathbf{0}}
\def\1{\mathbf{1}}

\def\A{{\bf A}}
\def\B{{\bf B}}
\def\C{{\bf C}}
\def\D{{\bf D}}
\def\E{{\bf E}}
\def\F{{\bf F}}
\def\G{{\bf G}}
\def\H{{\bf H}}
\def\I{{\bf I}}
\def\J{{\bf J}}
\def\K{{\bf K}}
\def\L{{\bf L}}
\def\M{{\bf M}}
\def\N{{\bf N}}
\def\O{{\bf O}}
\def\P{{\bf P}}
\def\Q{{\bf Q}}
\def\R{{\bf R}}
\def\S{{\bf S}}
\def\T{{\bf T}}
\def\U{{\bf U}}
\def\V{{\bf V}}
\def\W{{\bf W}}
\def\X{{\bf X}}
\def\Y{{\bf Y}}
\def\Z{{\bf Z}}

\def\a{{\bf a}}
\def\b{{\bf b}}
\def\c{{\bf c}}
\def\d{{\bf d}}
\def\e{{\bf e}}
\def\f{{\bf f}}
\def\g{{\bf g}}
\def\h{{\bf h}}
\def\i{{\bf i}}
\def\j{{\bf j}}
\def\k{{\bf k}}
\def\l{{\bf l}}
\def\m{{\bf m}}
\def\n{{\bf n}}
\def\o{{\bf o}}
\def\p{{\bf p}}
\def\q{{\bf q}}
\def\r{{\bf r}}
\def\s{{\bf s}}
\def\t{{\bf t}}
\def\u{{\bf u}}
\def\v{{\bf v}}
\def\w{{\bf w}}
\def\x{{\bf x}}
\def\y{{\bf y}}
\def\z{{\bf z}}

\def\balpha{\mbox{\boldmath{$\alpha$}}}
\def\bbeta{\mbox{\boldmath{$\beta$}}}
\def\bdelta{\mbox{\boldmath{$\delta$}}}
\def\bgamma{\mbox{\boldmath{$\gamma$}}}
\def\blambda{\mbox{\boldmath{$\lambda$}}}
\def\bsigma{\mbox{\boldmath{$\sigma$}}}
\def\btheta{\mbox{\boldmath{$\theta$}}}
\def\bomega{\mbox{\boldmath{$\omega$}}}
\def\bxi{\mbox{\boldmath{$\xi$}}}
\def\bnu{\mbox{\boldmath{$\nu$}}}                                  
\def\bphi{\mbox{\boldmath{$\phi$}}}
\def\bmu{\mbox{\boldmath{$\mu$}}}

\def\bDelta{\mbox{\boldmath{$\Delta$}}}
\def\bOmega{\mbox{\boldmath{$\Omega$}}}
\def\bPhi{\mbox{\boldmath{$\Phi$}}}
\def\bLambda{\mbox{\boldmath{$\Lambda$}}}
\def\bSigma{\mbox{\boldmath{$\Sigma$}}}
\def\bGamma{\mbox{\boldmath{$\Gamma$}}}
                                  
\newcommand{\myprob}[1]{\mathop{\mathbb{P}}_{#1}}

\newcommand{\myexp}[1]{\mathop{\mathbb{E}}_{#1}}

\newcommand{\mydelta}[1]{1_{#1}}

\newcommand{\myminimum}[1]{\mathop{\textrm{minimum}}_{#1}}
\newcommand{\mymaximum}[1]{\mathop{\textrm{maximum}}_{#1}}    
\newcommand{\mymin}[1]{\mathop{\textrm{minimize}}_{#1}}
\newcommand{\mymax}[1]{\mathop{\textrm{maximize}}_{#1}}
\newcommand{\mymins}[1]{\mathop{\textrm{min.}}_{#1}}
\newcommand{\mymaxs}[1]{\mathop{\textrm{max.}}_{#1}}  
\newcommand{\myargmin}[1]{\mathop{\textrm{argmin}}_{#1}} 
\newcommand{\myargmax}[1]{\mathop{\textrm{argmax}}_{#1}} 
\newcommand{\myst}{\textrm{s.t. }}

\newcommand{\denselist}{\itemsep -1pt}
\newcommand{\sparselist}{\itemsep 1pt}

\definecolor{pink}{rgb}{0.9,0.5,0.5}
\definecolor{purple}{rgb}{0.5, 0.4, 0.8}   
\definecolor{gray}{rgb}{0.3, 0.3, 0.3}
\definecolor{mygreen}{rgb}{0.2, 0.6, 0.2}

\newcommand{\cyan}[1]{\textcolor{cyan}{#1}}
\newcommand{\red}[1]{\textcolor{red}{#1}}  
\newcommand{\blue}[1]{\textcolor{blue}{#1}}
\newcommand{\magenta}[1]{\textcolor{magenta}{#1}}
\newcommand{\pink}[1]{\textcolor{pink}{#1}}
\newcommand{\green}[1]{\textcolor{green}{#1}} 
\newcommand{\gray}[1]{\textcolor{gray}{#1}}    
\newcommand{\mygreen}[1]{\textcolor{mygreen}{#1}}    
\newcommand{\purple}[1]{\textcolor{purple}{#1}}       

\definecolor{greena}{rgb}{0.4, 0.5, 0.1}
\newcommand{\greena}[1]{\textcolor{greena}{#1}}

\definecolor{bluea}{rgb}{0, 0.4, 0.6}
\newcommand{\bluea}[1]{\textcolor{bluea}{#1}}
\definecolor{reda}{rgb}{0.6, 0.2, 0.1}
\newcommand{\reda}[1]{\textcolor{reda}{#1}}

\def\changemargin#1#2{\list{}{\rightmargin#2\leftmargin#1}\item[]}
\let\endchangemargin=\endlist
                                               
\newcommand{\cm}[1]{}

\newcommand{\mhoai}[1]{{\color{blue}{[MH: #1]}}}

\newcommand{\mtodo}[1]{{\color{red}$\blacksquare$\textbf{[TODO: #1]}}}
\newcommand{\myheading}[1]{\vspace{1ex}\noindent \textbf{#1}}
\newcommand{\htimesw}[2]{\mbox{$#1$$\times$$#2$}}

\newif\ifshowsolution
\showsolutiontrue

\ifshowsolution  
\newcommand{\Solution}[2]{\paragraph{\bf $\bigstar $ SOLUTION:} {\sf #2} }
\newcommand{\Mistake}[2]{\paragraph{\bf $\blacksquare$ COMMON MISTAKE #1:} {\sf #2} \bigskip}
\else
\newcommand{\Solution}[2]{\vspace{#1}}
\fi

\newcommand{\truefalse}{
\begin{enumerate}
	\item True
	\item False
\end{enumerate}
}

\newcommand{\yesno}{
\begin{enumerate}
	\item Yes
	\item No
\end{enumerate}
}

\newcommand{\Sref}[1]{Sec.~\ref{#1}}
\newcommand{\Eref}[1]{Eq.~(\ref{#1})}
\newcommand{\Fref}[1]{Fig.~\ref{#1}}
\newcommand{\Tref}[1]{Table~\ref{#1}}

\definecolor{mygray}{gray}{.9}
\makeatletter
\newcommand{\thickhline}{%
    \noalign {\ifnum 0=`}\fi \hrule height 0.8pt
    \futurelet \reserved@a \@xhline
}

\newlength{\DepthReference}
\settodepth{\DepthReference}{g}%

\newlength{\HeightReference}
\settoheight{\HeightReference}{T}

\newlength{\Width}%

\newcommand{\MyColorBox}[2][red]%
{%
    \settowidth{\Width}{#2}%
    \colorbox{#1}%
    {%
        \raisebox{-\DepthReference}%
        {%
                \parbox[b][\HeightReference\DepthReference][c]{\Width}{\centering#2}%
        }%
    }%
}

\definecolor{mygray}{gray}{.9}
\makeatletter